\documentclass[11pt]{article}

\usepackage{acl}

\usepackage{times}
\usepackage{latexsym}
\usepackage[T1]{fontenc}
\usepackage[utf8]{inputenc}
\usepackage{microtype}
\usepackage{inconsolata, multirow}
\usepackage{booktabs}
\usepackage{array}
\usepackage{graphicx}
\usepackage{xcolor}
\usepackage{enumitem}
\usepackage{listings}
\usepackage{amsmath,amsfonts,bm}

\def\eqref#1{equation~\ref{#1}}
\def\1{\bm{1}}

\DeclareMathAlphabet{\mathsfit}{\encodingdefault}{\sfdefault}{m}{sl}
\SetMathAlphabet{\mathsfit}{bold}{\encodingdefault}{\sfdefault}{bx}{n}

\usepackage{cleveref}
\usepackage{placeins}
\usepackage{tabularx}
\usepackage[normalem]{ulem}


\lstdefinestyle{prompt}{
  language=,                % treat as plain text
  basicstyle=\ttfamily\scriptsize,
  breaklines=true,          % wrap long lines
  breakindent=0pt,
  columns=fullflexible,
  backgroundcolor=\color{black!3},
  frame=single,
  framerule=0.4pt,
  rulecolor=\color{black!40},
  xleftmargin=0pt,
  xrightmargin=0pt,
  aboveskip=4pt,
  belowskip=4pt,
}

\graphicspath{{figures/}}

\definecolor{darkblue}{rgb}{0, 0, 0.5}
\hypersetup{colorlinks=true, citecolor=darkblue, linkcolor=darkblue, urlcolor=darkblue}

\title{Knowledge Acquisition During Pre-training?\\ Large Language Models Learn Better With Auxiliary Views}

\author{
  \textbf{Joseph Lee},
  \textbf{Yidi Huang},
  \textbf{Dokyoon Kim},
  \textbf{Shu Yang}\thanks{Corresponding authors},
  \textbf{Li Shen}\footnotemark[\value{footnote}] \\
  University of Pennsylvania, Philadelphia, PA, USA \\
  \texttt{jiosephlee@gmail.com} \\
  \texttt{\{yidi.huang,dokyoon.kim,shu.yang,li.shen\}@pennmedicine.upenn.edu}
  \\[0.75ex]
  {\normalsize\rmfamily\upshape
    \href{https://huggingface.co/datasets/jiosephlee/auxiliary-views-knowledge-acquisition}{%
      \raisebox{-3pt}{\includegraphics[height=14pt]{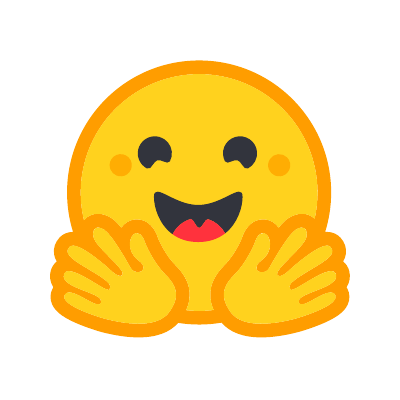}}
      Dataset}
    \qquad
    \href{https://github.com/jiosephlee/auxiliary-views-knowledge-acquisition}{%
      \raisebox{-3pt}{\includegraphics[height=14pt]{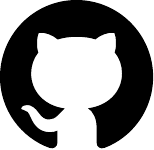}}
      Code}
  }
}

\begin{document}

\maketitle

\begin{abstract}
Gaps remain in our understanding of how large language models (LLMs) acquire knowledge during pre-training. We find that auxiliary views, conceptual reformulations, are key to this process. We design controlled experiments to isolate this. First, holding the token budget fixed, \textit{allocating tokens from paraphrases to auxiliary views} improves learning even for factual recall, which we use near-verbatim sentences from the original document to test. Second, the effectiveness of auxiliary views is \textit{not contingent} on the strength of the model that generates them. Third, we establish that paraphrasing is helpful but only at smaller batch sizes. Fourth, we identify types of knowledge that aid learning in the presence of prior knowledge gaps. Finally, we examine how these effects manifest mechanistically via \textit{layer-wise biases and compression}. Together, our findings suggest that auxiliary representations of knowledge are a key factor in the success of pre-training, and explain why data diversity matters. Further, our results provide clear recipes for synthetic data augmentation along with implications on the self-improvement in LLMs.
\end{abstract}

\section{Introduction}

Various aspects of data have been shown to be important in pre-training, including deduplication~\citep{raffel2020exploring, lee2021deduplicating, zhang2022opt}, filtering~\citep{weber2024redpajama, li2024datacomp}, coverage and depth~\citep{kandpal2023large}, quality~\citep{gunasekar2023textbooks, longpre2024pretrainer}, and diversity~\citep{chen2025revisiting, zhang2025harnessing}. However, these insights concern corpus characteristics, overlooking a fundamental question: \textit{how should knowledge be represented?}

The complexity of this question can vary. For atomic facts like biographical attributes, representation is trivial. But for domains such as biomedicine or law, where knowledge is multifaceted and interdependent, the right formulation is far less obvious. This is increasingly relevant as researchers adapt LLMs to specialized domains via continued pre-training~\citep{bai2025intern,sellergren2025medgemma,wang2025txgemma,luo2025large,colombo2024saullm, singhal2025toward}.
\begin{figure}[t]
    \centering
    \includegraphics[width=1.025\columnwidth]{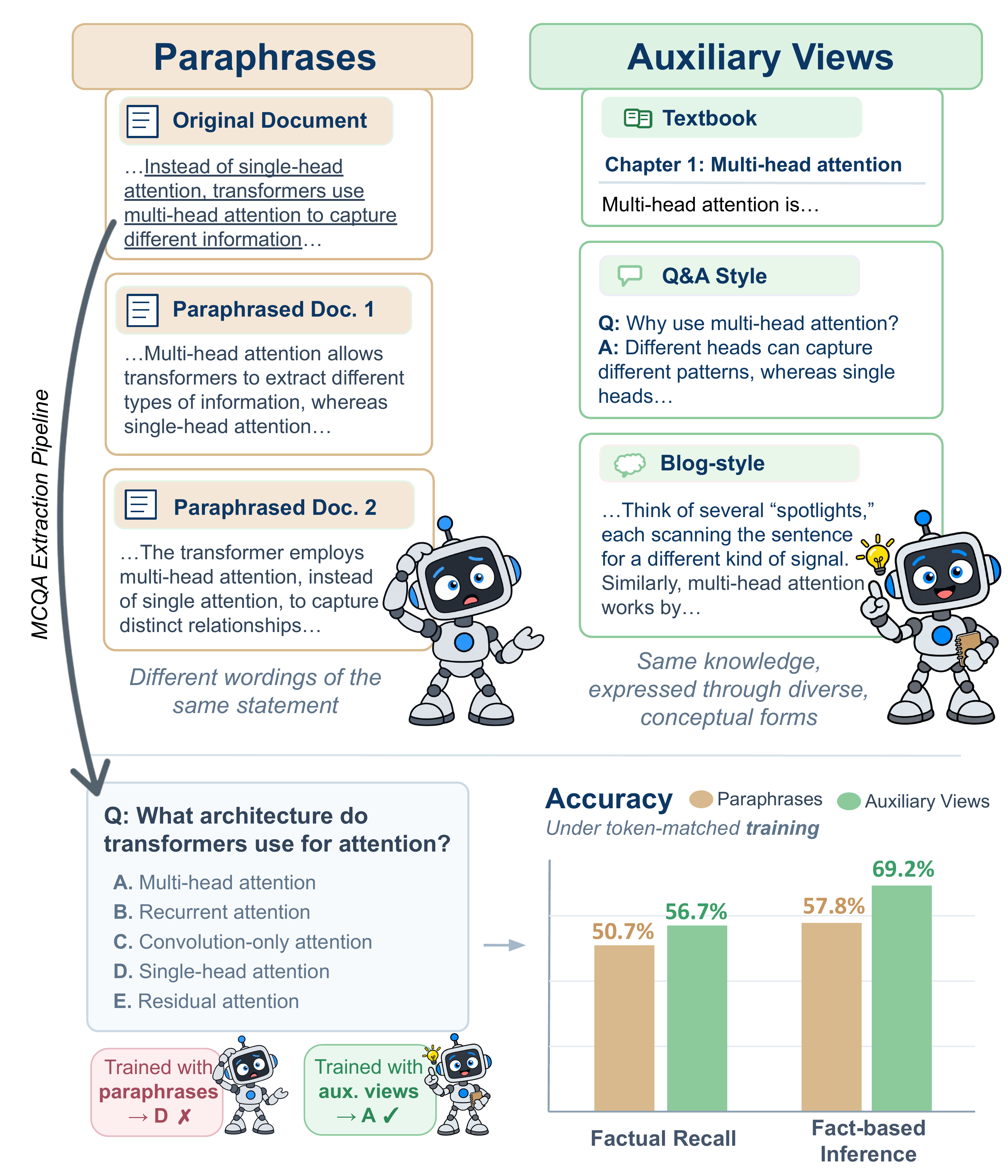}
\caption{\small{Under token-matched training, auxiliary views yield stronger knowledge acquisition than paraphrasing. The evaluation was designed by converting each knowledge-bearing sentence in the original document into MCQA.}}
  \label{fig1}
  \vspace{-6pt}
\end{figure}

Thus, for our experiments, we collect unseen arXiv papers, legal opinions, and medical cases to serve as self-contained domains. We continue pre-training on these texts and related formulations to study the acquisition of \textit{complex knowledge which builds upon the LLM's existing knowledge}. In this setting, we study two research questions:

\textbf{RQ 1:} How should knowledge be represented in natural language?

\textbf{RQ 2:} What surrounding knowledge should be represented alongside it?

Our experiments reveal four main findings. (1) Acquisition of new knowledge benefits from repetition, and paraphrasing further helps, but the \textbf{benefits of paraphrasing diminish at larger batch sizes}. (2) Holding the token budget fixed, \textbf{allocating tokens from document repetition to auxiliary views} improves understanding and \textbf{factual recall}, inducing a \textbf{distinct layer-wise bias and compression} in how knowledge is encoded. (3) Furthermore, the effectiveness of these auxiliary views is \textbf{not contingent on the strength of the teacher model} that generates them. (4) Lastly, contextual and prerequisite knowledge aid learning in the presence of \textbf{prior knowledge gaps}.

Together, these results lead us to a central conjecture: during pre-training, LLMs benefit from \textit{auxiliary views} (a web of explanations, analogies, and reformulations that humans generate as they learn and teach each other) and acquire a more generalizable encoding of knowledge. Conceptually, not just linguistically, diverse views of the same knowledge improve learning: \textit{broader conceptual understanding facilitates the memorization of specific facts better than brute memorization}. Moreover, this ability emerges more strongly with scale: \textit{larger models learn better by integrating diverse views more effectively}, encoding them with greater parameter efficiency and redistributing learning toward the middle and final layers.

Our findings provide a more operational account of what ``diverse'' data can mean in pre-training. Rather than treating diversity solely as a corpus-level property, diversity can be constructed \textit{around individual knowledge}, through complementary views that help models form richer and more efficient representations. This offers principles for synthetic pre-training, especially in low-data domains, and explains why ``diverse'' data is helpful.

\section{Related Works}
\label{related_works}

\textbf{Pre-training.} Model size and data scale are the primary determinants of LLM capabilities~\citep{brown2020language, kaplan2020scaling, carlini2022quantifying, tirumala2022memorization, hoffmann2022training}. The large-scale nature of pre-training has made the exact role of data difficult to understand. Existing studies, varying in corpora and models, often yield conflicting findings. Even the role of data repetition remains debated~\citep{lee2021deduplicating, taylor2022galactica, hernandez2022scaling, xue2023repeat, muennighoff2023scaling}.

\textbf{Knowledge Acquisition during Pre-training.} We therefore build upon previous efforts that study how pre-training instills knowledge in a controlled manner. \citet{allen2024physics} show that paraphrased augmentation increases the memorization of biographical facts from 9.7\% to 96.6\%. They find that inserting QAs about facts during pre-training increases memorization of held-out facts, whereas inserting QAs during instruction tuning does not. This strongly suggests the importance of data formulation in pre-training.

Further, \citet{chang2024large} intermittently injects fictional facts during pre-training, observing that knowledge is acquired incrementally upon each exposure and subject to decay, suggesting a gradual, not emergent, learning process. Confounding factors are controlled to clearly isolate the effect.

A limitation of these prior works is that \citet{chang2024large} and \citet{allen2024physics} only use biographical facts; they also provide conflicting views on the benefits of paraphrasing. We focus on complex knowledge and offer an explanation for these reported differences.

\textbf{Domain Adaptation via Continued Pre-training.}
Teaching large language models (LLMs) a specific set of new knowledge via continued pre-training is difficult
\citep{wang2021can, jang2021towards, hu2023meta,ovadia2024fine, hoffbauer-etal-2024-knowledge}. For instance, a 70B model continually pretrained on Wiki-style documents, despite sophisticated augmentation, recalls 62.7\% of facts~\citep{jiang2024instruction}. In practice, continued pre-training is adopted to varying degrees. For example, MedGemma~\citep{sellergren2025medgemma} and TxGemma~\citep{wang2025txgemma} do not perform any text-based continued pre-training, while Intern-S1~\citep{bai2025intern} does so for 5 trillion tokens. Thus, our study aims to provide practical takeaways that can make continued pre-training more reliable.

\begin{table}[t]
\centering
{\footnotesize
\begin{tabularx}{\linewidth}{@{} >{\raggedright\arraybackslash}X @{}}
\toprule
\textbf{Blog.} Direct Preference Optimization (DPO) offers a neat change of perspective. Instead of treating reward modeling and policy optimization as separate steps, DPO re-parameterizes the reward class so [...] has a closed form. That lets you fit the policy directly to human comparisons with a simple classification loss --- no RL required. \tabularnewline
\midrule
\textbf{Stack Exchange.} \textbf{Q:} How exactly does the partition function cancel out in the DPO derivation? They say the partition function term $\beta \log Z(x)$ cancels --- can someone show the algebraic steps explicitly? \textbf{A:} Short answer: because the partition function term in the reparameterization depends only on the prompt $x$, it is identical for both completions and therefore subtracts out when you form the [...]\tabularnewline
\midrule
\textbf{Textbook.} The partition function term $\beta\log Z(x)$ depends only on $x$, so it cancels when forming differences. Preference models like Bradley--Terry depend only on reward differences, not on absolute reward values. Therefore the unknown normalization that made reward recovery hard is irrelevant to the likelihood of observed pairwise comparisons. This is precisely why we can reparameterize rewards in terms of [...] \tabularnewline
\bottomrule
\end{tabularx}
}
\caption{\small{We rewrite each document into three genres while preserving the underlying knowledge. The excerpts above illustrate an accessible, narrative \textit{blog}; a question-and-answer \textit{Stack Exchange} entry; and a formal \textit{textbook} exposition.}}
\label{tab1}
\vspace{-5mm}
\end{table}

\begin{table*}[t]
\centering
{\footnotesize
\resizebox{0.98\textwidth}{!}{
\begin{tabular}{@{} >{\raggedright\arraybackslash}p{3cm} >{\raggedright\arraybackslash}p{12cm} @{}}
\toprule
\textbf{Original sentence} & The \textit{added constraint is important}, as it \textit{prevents the model from} deviating too far from the distribution [...], as well as \textit{maintaining the generation diversity} and [...] \\
\midrule
\textbf{Factual probe} & In the paper ``Direct Preference Optimization [...],'' the authors state that the \textit{added constraint} in the reinforcement learning objective not only \textit{prevents the model from} deviating too far from the distribution on which the reward model is accurate, but also \textit{maintains} \textbf{the generation diversity}. \\
\midrule
\textbf{Support sentences} & [...] still \textit{expensive to estimate the partition function} $Z(x)$ [...] the Bradley-Terry model \textit{depends only on the difference of rewards} [...] Substituting the reparameterization [...] into the preference model, \textit{the partition function cancels}, and we can express the human preference probability in terms of only the optimal policy $\pi^*$ and reference policy [...] \\
\midrule
\textbf{Inference probe} & According to the paper ``Direct Preference Optimization [...],'' the expensive quantity from the optimal-policy form that becomes unnecessary to estimate once DPO rewrites the Bradley--Terry preference model in terms of policies is \textbf{the partition function}. \\
\bottomrule
\end{tabular}
}}
\caption{\small{The \textit{factual} probe extracts questions from knowledge-bearing sentences, converts them into statements with answers at the end, and adds context. The \textit{inference} probe combines knowledge from one or more support sentences to infer information not explicitly stated. Here, it integrates the facts that $Z(x)$ is expensive to estimate, that the Bradley--Terry model depends only on reward differences, and that the reparameterization causes $Z(x)$ to cancel, thereby identifying the partition function as the quantity that no longer requires estimation. The \textbf{target span} of each probe is bolded.}}
\label{tab2}
\vspace{-5mm}
\end{table*}

\section{Experimental Setup}
\label{experiment}

\textbf{Problem Formulation.}
The pre-training objective for an autoregressive LM, $f$ parameterized by $\theta$, is next-token prediction.
Given a corpus $C$ of documents ${d_1,\dots,d_M}$, each a sequence of tokens $(t_{m,1}, \dots, t_{m,n_m})$,
we minimize the loss function:

\begin{equation}
L(\theta) = -\sum_{m=1}^{M} \sum_{i=1}^{n_m} \log P(t_{m,i} \mid t_{m,<i}; \theta)
\end{equation}

We denote knowledge abstractly as $K$, accessible through the sub-corpus $\mathcal{C}_K = \{\, d_i \in \mathcal{C} \mid d_i \text{ contains information about } K \,\}$. Because $K$ is acquired only through next-token prediction over $\mathcal{C}_K$, the selection of tokens should matter. This motivates our central question: how should knowledge be represented in text, or specifically how should we construct documents $d \in \mathcal{C}_K$ to better learn $K$?

\textbf{Observation.} Knowledge is rarely represented once in a corpus. Consider ``Attention Is All You Need''~\citep{vaswani2017attention}. This document is followed by a proliferation of others conveying the same knowledge: tutorials, blogs with intuitive explanations, and forums answering questions. Such texts are \textit{natural byproducts of society's collective effort to communicate and process knowledge}.

We term such manifestations of the same knowledge \textbf{auxiliary views}. We make this distinction because human-generated reformulations are never mere paraphrases: they discuss knowledge in diverse contexts (e.g., Transformers' advantages over other architectures) and forms (e.g., blogs, forums), collectively providing a more \textit{complete, contextualized} picture of $K$. We suspect pre-training is replete with these views, explaining its effectiveness beyond vague axes like quality or diversity. In our study, we frame paraphrasing as linguistic variation of a single view and set it as the control.

We also investigate knowledge represented around $K$. To acquire $K$, a model may need prior knowledge it does not yet possess. We examine two types: \textbf{contextual knowledge}, which $K$ references, and \textbf{prerequisite knowledge}, foundational concepts that $K$ presupposes.

\subsection{Dataset}

\textbf{Documents.}
We collect 36 documents across three domains to serve as $K$: twelve computer science papers from arXiv, twelve legal opinions from U.S. federal appellate courts, and twelve medical case reports from PubMed Central. To prevent leakage from OLMo-2's pre-training corpus, we select documents published after the cutoff date and verify their absence through the Infini-gram API~\citep{liu2024infini}. Additional details are in Appendix~\ref{Appendix:preprocess}.

\textbf{Views.} For each document, we synthetically generate auxiliary views via LLMs. Inspired by our earlier observation and prior work~\citep{gunasekar2023textbooks, allen2024physics, jiang2024instruction}, we construct textbooks, Stack Exchange--style Q\&A, and blogs (Table~\ref{tab1}). Prerequisite knowledge is also generated as textbooks. Contextual knowledge is collected as cited arXiv papers and cited judicial opinions; medical case reports lack a citation structure, so we omit this domain. Paraphrases are generated with GPT-4.1, while auxiliary and prerequisite views are generated with GPT-5-mini. Details on data collection, preprocessing, and prompts are in Appendix~\ref{Appendix:preprocess}.

\textbf{Probes.} To measure learning, we adopt LAMA-style probes~\citep{petroni-etal-2019-language,jiang-etal-2020-know, zhong2021factual}, where the task is to predict a sentence's final word or, following \citet{chang2024large}, a multi-word target. We design two probe types (Table~\ref{tab2}): factual probes, which measure recall of information stated explicitly in the text, and inference probes, which require combining one or more facts to infer information not stated.

To construct factual probes, we filter for knowledge-bearing sentences, generate QA pairs from each, and convert them into self-contained cloze statements; we automate this with GPT-5.4, having verified that it adequately understands the documents. The pipeline yields 6,435 factual and 430 inference probes, along with multiple-choice variants (4,515 factual and 322 inference MCQs), exceeding the scale of prior work~\citep{chang2024large}. We manually validate 200 probes of each type. More pipeline details are in Appendix~\ref{Appendix:Probes}.

\subsection{Training Setup}

We use the base OLMo-2 models (1B, 7B, 13B, and 32B)~\citep{olmo20242} for training. Following \citet{chang2024large}, we employ single-batch knowledge injection: the documents fit in one forward pass, and the rest of the batch is filled with general data. We perform $N=100$ injections in our main experiments, one in each training step. In \textit{Source}, the original document is injected every batch. In \textit{Para.\ M}, we cycle through the original document and its $M$ paraphrases. In \textit{Para. M + Aux.}, auxiliary views are injected alongside the paraphrased documents: a blog, textbook chapter, and Stack Exchange Q\&A in every batch. In \textit{Aux. only}, only auxiliary views are inserted. We make sure to token-match all conditions: since \textit{Para. M + Aux.} introduces additional knowledge-bearing tokens, we upsample the chunks in \textit{Source} and \textit{Para.\ M} so that every condition sees the same number of tokens pertaining to $K$. Consequently, \textit{Para. M + Aux.} contains relatively fewer repetitions of the main document. Hyperparameters are in Appendix~\ref{append:hyperp}.

\textbf{General Data.} For general data replay, we stream tokens from the DCLM subset~\citep{li2024datacomp} of OLMo-2's pre-training corpus.

\textbf{Metrics.} We evaluate the model's performance on our probes throughout training. Metrics are reported at the final training step unless stated otherwise.
\begin{itemize}[itemsep=0pt, topsep=0pt, parsep=0pt]
    \item \textbf{Log-Prob.}: For cloze probes, we measure the joint log-probability of the target span.
    \item \textbf{Target Rank}: We measure the vocabulary rank of the correct token under teacher forcing. For multi-token targets, we report the worst rank across the span; lower is better.
    \item \textbf{Multiple-Choice Accuracy}: We use 5-shot prompting with general-knowledge examples and constrained decoding.
\end{itemize}

\section{Large Language Models Learn Better with Auxiliary Views}

\begin{figure}[h]
    \centering
    \includegraphics[width=\columnwidth]{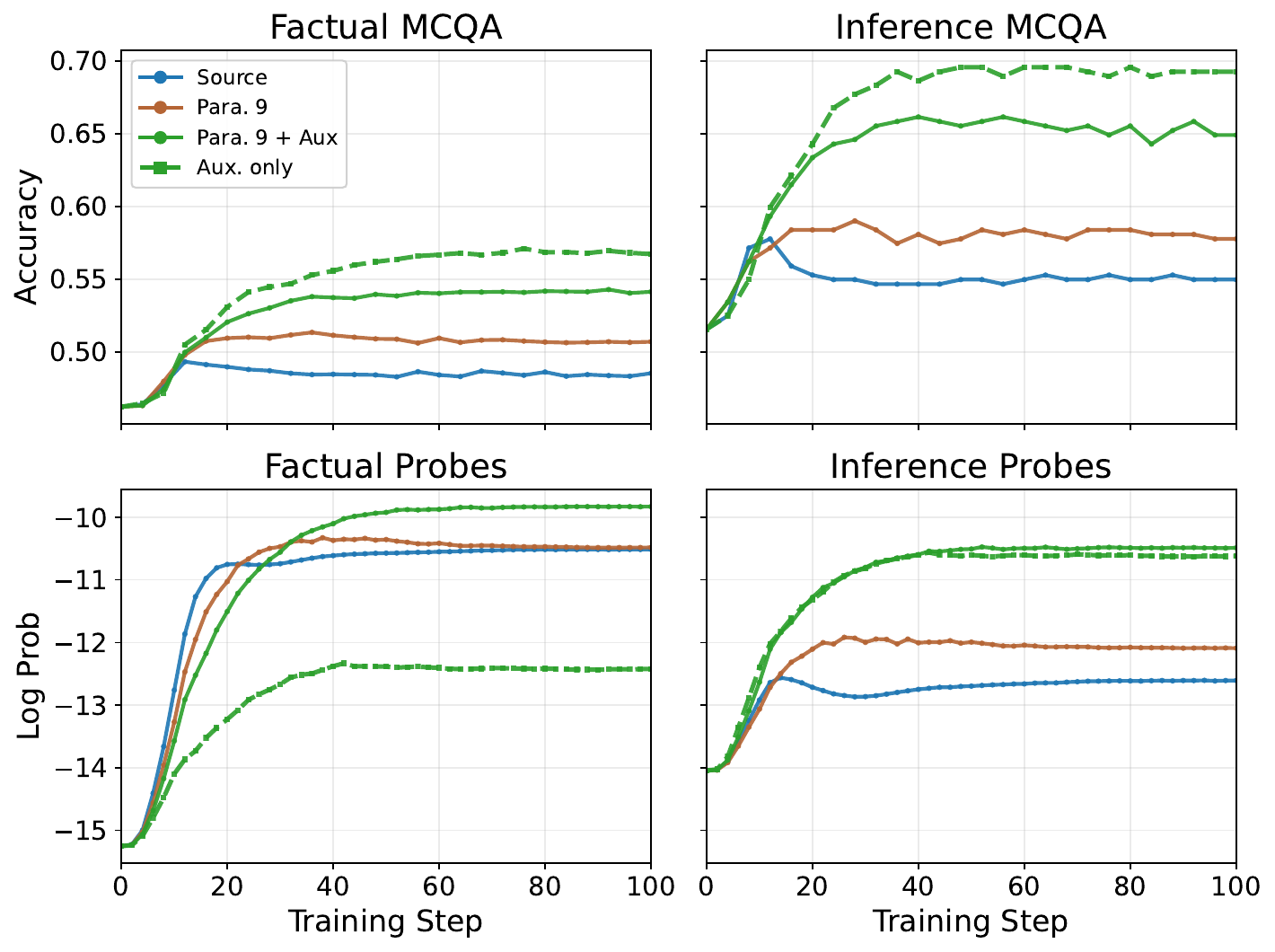}
\caption{{Four token-matched conditions on OLMo-2-32B. \textit{Aux. only} training yields the highest factual and inference MCQA accuracy, while \textit{Para. 9 + Aux.} yields the highest cloze log probabilities.}}
  \label{fig2:probe_32b}
\end{figure}

\begin{figure*}[t]
    \centering
    \includegraphics[width=\textwidth]{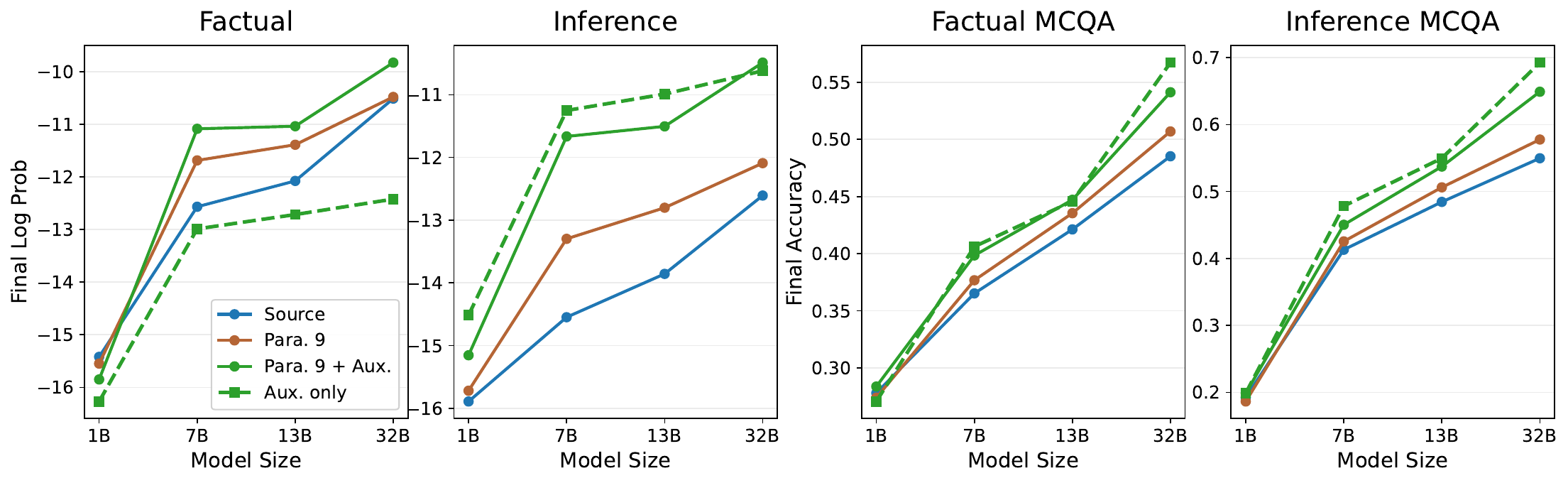}
\caption{Knowledge acquisition across model sizes under a fixed token budget. Auxiliary views (green) improve both factual recall and inference over \textit{Source} (blue) and \textit{Para. 9} (orange); this advantage grows with model size. The only exception is \textit{Aux. only}, which performs worse on the factual probes.}
    \label{views}
\end{figure*}

\subsection{The Benefit of Auxiliary Views}

\textbf{Better Knowledge Acquisition.} In Figure~\ref{fig2:probe_32b}, even under a matched token budget, allocating tokens to auxiliary views substantially improves performance on both factual recall and fact-based inference. For \textit{Para. 9 + Aux.}, this pattern holds across all of our metrics, including log probability, MCQA, and target rank (Figure~\ref{views_target_rank}). At the final step, \textit{Para. 9 + Aux.} outperforms \textit{Para.\ 9} and \textit{Source} on all four metrics, though \textit{Source} learns faster on factual cloze probes early in training. Replacing the entire knowledge-bearing budget with auxiliary views (\textit{Aux. only}) improves performance on both MCQA tasks further, while \textit{Para. 9 + Aux.} retains higher cloze log probabilities. We attribute these gains to auxiliary views as all other factors, including data ordering, are held constant.

The improvement on inference is intuitive. Although source documents contain sufficient information to answer these probes, they offer only a single perspective. As Table~\ref{tab1} illustrates, auxiliary views reformulate the knowledge in different styles and contexts, offering diverse perspectives on the material. Learning from these views should lead to a more generalizable understanding, whose benefits naturally emerge on inference probes.

\textbf{Generalization to Factual Recall.} Counterintuitively, auxiliary views also improve \textit{factual} recall, even though every factual target is a verbatim phrase from the source. One would expect diverting tokens away from source to \textit{hinder} direct memorization, not \textit{help it}. Auxiliary views surprisingly encourage the model to encode knowledge in a way that supports more effective factual recall; we return to this in Section~\ref{sec:mechanistic} by examining how this effect emerges mechanistically. Drawing parallels to human learning may offer a fruitful direction for future inquiry~\citep{spiro2017remembering}.

\textbf{Model Size Effects.} The benefit of auxiliary views further \textit{emerges with scale}. As shown in Figure~\ref{views}, the 1B model derives little advantage from them, while an advantage appears at 7B and grows with size. This trend is largely consistent across all domains (Figure~\ref{views_by_domain}). We discuss this further in Section~\ref{sec:mechanistic} in which larger models appear to unlock an advantage from auxiliary views that smaller models cannot.

\textbf{Which Auxiliary View?} We again token-match every condition. Textbooks, blogs, and Stack Exchange Q\&A all perform similarly, with a slight benefit from mixing view types on factual recall (Table~\ref{tab:view-family}). Whether and how each view facilitates learning differently warrants further study.

\textbf{Measuring Lexical Bias.} A possible confounding factor is that the synthetically generated auxiliary views leak our probes, especially for inference probes where the target span, unlike factual probes, need not appear in the source. We address the question of \textit{lexical} bias, whereby the model might favor particular phrases since auxiliary views and probes were produced by related model families (GPT-5-mini and GPT-5.4, respectively). We measure how often each probe's target span occurs in the source, paraphrases, and auxiliary views.

Table~\ref{tab:target-occurrence-all} reports both frequency and coverage. Across all metrics, auxiliary views contain the target spans \textit{less} frequently and with \textit{lower} coverage than either the source or paraphrases. This strengthens our findings: auxiliary views perform better on the probes despite stating the targets less often.
% Requires \usepackage{booktabs,multirow} in the preamble.
\begin{table}[!htbp]
\centering
\scriptsize
\setlength{\tabcolsep}{2.5pt}
\resizebox{\columnwidth}{!}{
\begin{tabular}{@{}lcccc@{}}
\toprule
\textbf{Model} & \textbf{Fact.} & \textbf{Fact.} & \textbf{Inf.} & \textbf{Inf.} \\
 & $\boldsymbol{\Delta}$ \textbf{log prob.} & $\boldsymbol{\Delta}$ \textbf{MCQA (pp)} & $\boldsymbol{\Delta}$ \textbf{log prob.} & $\boldsymbol{\Delta}$ \textbf{MCQA (pp)} \\
\midrule
OLMo-3 7B & -0.15 & +1.53 & +0.91 & +4.04 \\
Qwen-2.5 7B & +2.29 & +3.26 & +3.50 & +0.31 \\
Qwen-3 8B Base & +0.49 & +1.26 & +1.61 & +4.35 \\
Llama-3.1 8B & +1.35 & +3.08 & +2.63 & +5.90 \\
Gemma-3 12B & +1.00 & +1.22 & +2.05 & +1.24 \\
\bottomrule
\end{tabular}
}
\caption{The benefit of auxiliary views generalizes to other language models. $\boldsymbol{\Delta} = \textit{Para. 9 + Aux.} - \textit{Para. 9}$. MCQA changes are in absolute percentage points (pp).}
\label{tab:backbone-generalization}
\end{table}

\textbf{Controlling for Upsampling.}
\label{sec:upsampling-control}
To test whether the density of duplicates introduced by token-matched upsampling of \textit{Source} and \textit{Para.\ 9} explains the gap, we reduce the degree of upsampling. In the half-allocation regime, we halve the auxiliary-token allocation and its matching upsampling, which yields 32.07M knowledge-bearing tokens compared to 41.12M at full allocation. In the no-upsampling regime, \textit{Source} and \textit{Para.\ 9} receive no upsampling; auxiliary views instead replace the document budget. Still, auxiliary views retain their advantage over \textit{Para.\ 9}, and \textit{Aux. only} yields the highest accuracy (Table~\ref{tab:upsampling-control}).

\begin{table}[h]
\centering
\scriptsize
\setlength{\tabcolsep}{3pt}
\begin{tabular}{@{}llcc@{}}
\toprule
\textbf{Matching} & \textbf{Condition} & \textbf{Factual} & \textbf{Inference} \\
\textbf{regime} & & \textbf{MCQA} & \textbf{MCQA} \\
\midrule
\multirow{4}{*}{Half aux. allocation} & Source & 0.399 & 0.435 \\
 & Para.\ 9 & 0.403 & 0.448 \\
 & \textit{Para. 9 + Aux.} & 0.407 & 0.459 \\
 & Aux. only & \textbf{0.413} & \textbf{0.492} \\
\midrule
\multirow{4}{*}{No upsampling} & Source & 0.380 & 0.421 \\
 & Para.\ 9 & 0.389 & 0.415 \\
 & \textit{Para. 9 + Aux.} & 0.392 & 0.424 \\
 & Aux. only & \textbf{0.399} & \textbf{0.465} \\
\bottomrule
\end{tabular}
\caption{\textbf{Token-matching upsampling.} Peak MCQA accuracy when upsampling is halved or removed. Auxiliary views retain their advantage, and \textit{Source} never improves with less duplication. We use the same training setup with OLMo-2 7B over 100 steps.}
\label{tab:upsampling-control}
\end{table}

\subsection{Broader Generalization}
\label{sec:broader-generalization}

We examine pre-training-faithful continuation, human-written views, and five additional models.

\textbf{A Pre-training-Faithful Setting.} Our experiments so far use continued pre-training with a new learning-rate schedule and a smaller batch size. These choices may affect whether our findings generalize to the original pre-training regime. We therefore examine a pre-training-faithful setup (Table~\ref{tab:proper-pretraining}; full setup details in Appendix~\ref{append:hyperp}). The benefits of auxiliary views generalize here.

\begin{table}[h]
\centering
\scriptsize
\setlength{\tabcolsep}{2.5pt}
\resizebox{\columnwidth}{!}{
\begin{tabular}{@{}lcccc@{}}
\toprule
\textbf{Condition} & \textbf{Fact.} & \textbf{Fact.} & \textbf{Inf.} & \textbf{Inf.} \\
 & \textbf{log prob.} & \textbf{MCQA} & \textbf{log prob.} & \textbf{MCQA} \\
\midrule
Pretrained model & -16.30 & 0.343 & -14.79 & 0.413 \\
Source  & -10.00 & 0.372 & -12.85 & 0.421 \\
Para.\ 9  & -10.33 & 0.375 & -12.53 & 0.417 \\
\textit{Para. 9 + Aux.} & \textbf{-9.56} & 0.403 & -10.82 & 0.492 \\
Aux. only & -11.99 & \textbf{0.423} & \textbf{-10.62} & \textbf{0.509} \\
\bottomrule
\end{tabular}
}
\caption{\textbf{Pre-training-Faithful Setting.}  Final metrics after resuming OLMo-2 7B from step 925{,}000 for 100 steps with its optimizer state, original data stream and schedule, and global batch size 1{,}024.}
\label{tab:proper-pretraining}
\end{table}

\begin{figure}[h]
    \centering
    \includegraphics[width=0.9\columnwidth]{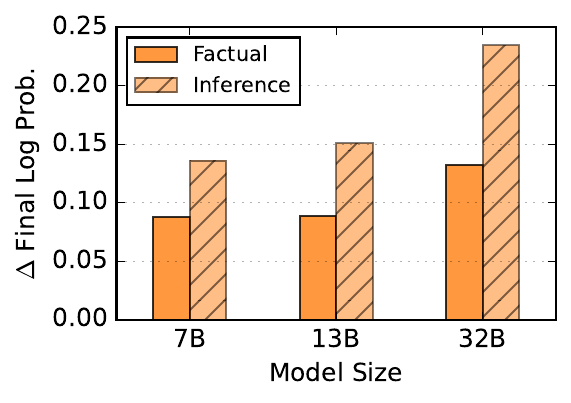}
    \caption{With human-written auxiliary views collected from the open web, the benefit grows with model size, consistent with our synthetic results. \(\Delta\) final log prob.\ is the difference between training with human auxiliary views and the \textit{Para.\ 9} condition. Results are averaged over two documents.}
    \label{fig:human}
\end{figure}

\textbf{Human Auxiliary Views.} To test whether our findings depend on synthetic text, we repeat the experiment with human-written auxiliary views collected from the open web. The benefit again grows with model size (Figure~\ref{fig:human}), mirroring the synthetic-view results. This experiment covers only two documents, so we treat it as suggestive rather than conclusive; nonetheless, it provides initial evidence that the effect is not merely an artifact of clean, synthetic text.

\textbf{Beyond OLMo-2.} We repeat the main experiment on the base models of OLMo-3 7B, Qwen-2.5 7B, Qwen-3 8B, Llama-3.1 8B, and Gemma-3 12B. Adding auxiliary views improves factual recall and inference for nearly all five language models (Table~\ref{tab:backbone-generalization}), except for OLMo-3, where factual cloze log probability decreases slightly.

\subsection{Mechanistic Signatures of Auxiliary Views}
\label{sec:mechanistic}

\begin{figure}[h]
    \centering
    \includegraphics[width=\columnwidth]{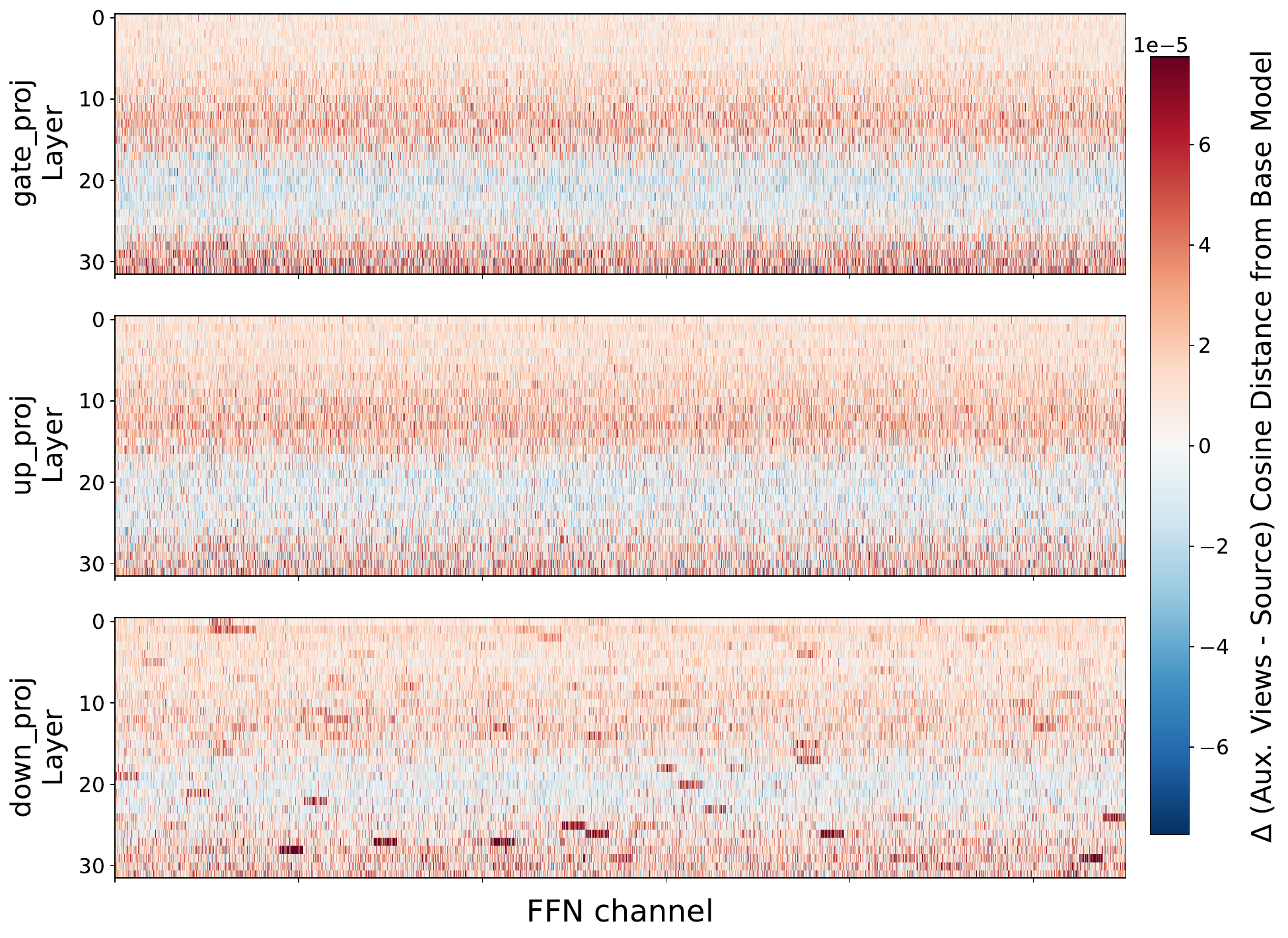}
    \caption{Per-channel difference (\textit{Para. 9 + Aux.} minus \textit{Source}) in cosine distance from the base model across layers and FFN channels (gate, up, and down projections). Red indicates channels that auxiliary views move more than source; blue indicates less movement. A negative band around layers 16--24 shows that auxiliary views change the upper-middle layers \textit{less} than source while affecting the middle and final layers \textit{more}.}
    \label{fig:mech_heatmap}
\end{figure}

Having established that auxiliary views improve knowledge acquisition, we ask how this phenomenon emerges mechanistically. We compare each trained model's feed-forward network (FFN) weights against the base model along two axes: the \textit{magnitude} of change (relative delta norm and cosine distance) and its \textit{concentration} across channels (Gini coefficient), measured per layer and over training (Figures~\ref{fig:mech_layer} and~\ref{fig:mech_steps}).

\textbf{Why MLP Channels?} We focus our analysis on the feed-forward (MLP) layers for two reasons. First, they account for the majority of a transformer's parameters, making them the natural locus for studying where knowledge is written during training. Second, they have comparatively clean interpretations: \citet{geva2021transformer} show that individual channels in FFN layers operate as key--value memories. This channel-level view lets us read parameter change as movement in the model's memory rather than as an opaque aggregate.

\textbf{Compression: Learning More by Changing Less.} Based on Figure~\ref{fig:mech_steps}, paraphrasing induces the \textit{largest} parameter movement of the three conditions, in both norm and cosine distance, yet adding auxiliary views \textit{reduces} this magnitude, though it remains greater than in source-only training. We interpret this as evidence that the two conditions learn differently. Paraphrasing supplies many surface-level variants of a single view, and the model appears to expend parameter change absorbing this lexical variation. Auxiliary views instead supply the knowledge in genuinely distinct framings, which the model can apparently integrate with less weight movement. This is consistent with our conjecture that auxiliary views enable a more generalizable encoding of knowledge: forming a more general, reusable representation requires overwriting fewer parameters than memorizing surface forms. Magnitude of change is not the same as quality of learning.

\textbf{Layer-wise Biases.} Parameter change is not uniform across layers. All conditions concentrate change in the middle and final layers, with a pronounced dip in the upper-middle layers (Figure~\ref{fig:mech_layer}). Auxiliary views accentuate this structure. Relative to training with \textit{Source}, training with auxiliary views changes the middle and final layers \textit{more}, but \textit{changes less} in the upper-middle band (layers \(\sim\)16--24). Figure~\ref{fig:mech_heatmap} makes this bias visible, and it is consistent across nearly every channel (row) in the FFN. Auxiliary views do not move more weight everywhere; they redistribute \textit{where} learning occurs.

\textbf{Model Size Effects.} At 1B, where auxiliary views confer little benefit (Figure~\ref{views}), the corresponding per-channel difference is almost uniformly negative or zero across all layers and projections (Figure~\ref{fig:mech_heatmap_1b}), lacking the layer-wise biases seen at 7B. The mechanistic signature of auxiliary views therefore emerges only at scale. This suggests that \textbf{larger models encode auxiliary views differently}, enabling greater generalization.

\begin{figure*}[t]
    \centering
    \includegraphics[width=\textwidth]{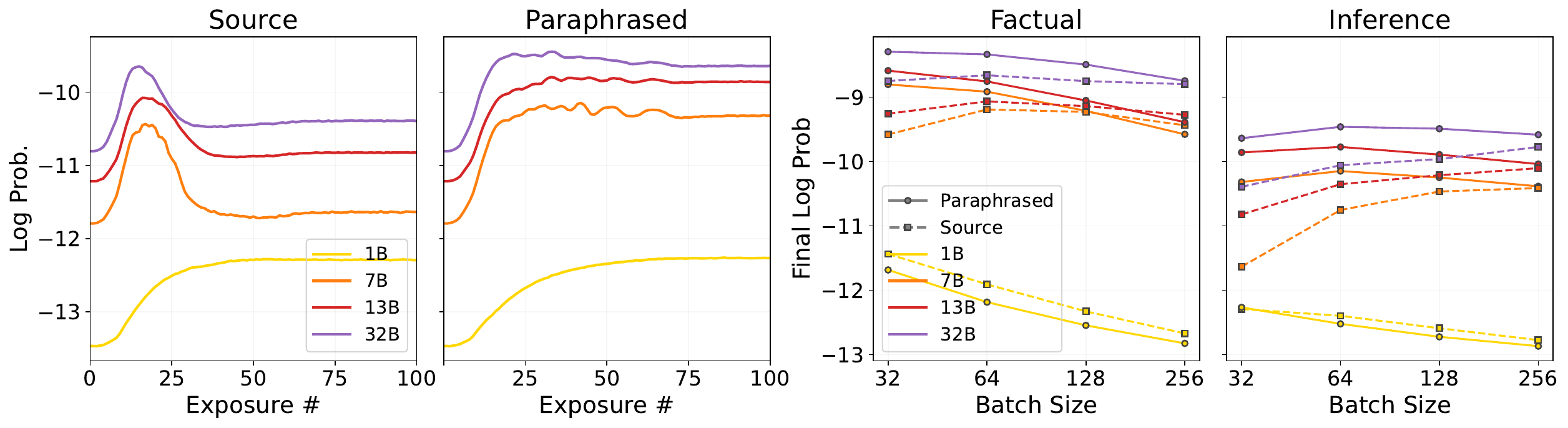}
    \caption{\textbf{(Left)} Inference-probe log probability while training OLMo-2-7B with a batch size of 64 on \textit{Source} versus \textit{Para.~9}. \textbf{(Right)} Final probe log probabilities for \textit{Source} versus \textit{Para.~9} across model sizes and batch sizes. This experiment uses six documents to examine dynamics at smaller batch sizes.}
    \label{fig:batch_size}
\end{figure*}

\section{Auxiliary Views Do Not Require a Strong Teacher}
\label{sec:generator-ablation}

Auxiliary views inherently require a teacher: someone who understands the material and can communicate it to others. Naturally, our results could hinge on the generator's strength, although our auxiliary text is generated by a relatively weak model, \textsc{gpt-5-mini}. To examine the influence of the teacher's capabilities, we regenerate auxiliary views using eleven generator configurations spanning multiple model families, reasoning-effort levels, and sizes, while holding the training schedule and token budget fixed. Because generators that produce less auxiliary-view text require more repetition to match the token budget and may therefore overfit, we compare each at its peak over the 100-step training window.

Downstream factual accuracy is remarkably stable across generators (0.385--0.405, all above the \textit{Para.\ 9} baseline of 0.377; Table~\ref{tab:generator-ablation}). It is uncorrelated with generator size (Pearson $r=-0.11$, $n=6$), as further corroborated by scaling within a model family (gpt-oss-20B$\rightarrow$120B: 0.405$\rightarrow$0.394; Gemma-4 12B$\rightarrow$31B: 0.392$\rightarrow$0.386), and likewise uncorrelated with the generator's own factual accuracy ($r=-0.18$, $n=10$). Notably, gpt-oss-20B has the least measured domain knowledge, yet its views teach best.

However, the volume of generated view text, which does not measure teacher strength and can be controlled through prompting, does correlate with downstream accuracy ($r=+0.68$, $n=11$, $p=0.020$). Varying the reasoning effort of the same generator likewise leaves downstream performance essentially unchanged, even though it changes the generator's own accuracy. Together, these results suggest that auxiliary views function as general data augmentation rather than as distillation from a strong teacher. The generator need only reformulate the provided text; stronger models do not necessarily perform better at this task, even in complex domains.

\section{When Does Paraphrasing Help?}
\label{paraphrasing}

Paraphrasing is known to aid knowledge acquisition~\citep{ovadia2024fine}, but prior reports conflict on its effectiveness. We revisit this question to show that its benefit is conditional, and in doing so reconcile these reports.

\textbf{Paraphrasing prevents collapse.} Repeated exposure improves performance only up to $\sim$20 exposures, after which \textit{Source} saturates and then sharply degrades as the model overfits (Figure~\ref{fig:batch_size}, left). \textit{Para.\ 9} prevents this collapse, sustaining improvement through $\sim$40 exposures. As with auxiliary views, this benefit emerges only at 7B and above; at 1B, paraphrasing is slightly harmful.

\textbf{The benefit depends on batch size.} However, we see that this advantage depends on batch size, likely because larger batches mix in more general data per step and this similarly suppresses overfitting. As batch size grows, \textit{Source} becomes more stable and nearly matches \textit{Para.\ 9} by batch size 256, while \textit{Para.\ 9} stays roughly flat (Figure~\ref{fig:batch_size}, rightmost). Paraphrasing's inference gains thus come from preventing a degeneration that large batches also prevent; together, the two interventions become redundant.

Factual acquisition behaves differently. At small batch sizes, paraphrasing improves factual learning at 7B and above, whereas \textit{Source} sees almost no gain from larger batches (aside from a small improvement at batch size 64 for 7B). We attribute this advantage to how paraphrasing semantically varies the facts: although factual probes are drawn explicitly from source sentences, they are recast as self-contained atomic statements, so \textit{Source}'s verbatim memorization transfers less well. The advantage shrinks at larger batches and reverses at batch size 256 (7B and 13B), likely because the increased general-data mixing dilutes the gradient signal from the paraphrased text.

{Table~\ref{tab:proper-pretraining} supports this general interpretation at a batch size of 1{,}024: paraphrasing provides no consistent improvement over \textit{Source}, while auxiliary views retain a substantial advantage.}

\textbf{Reconciling Prior Works.} This batch-size dependence reconciles conflicting prior work: \citet{chang2024large}, at batch size 2048 with 2048-token chunks, reports \textit{degraded} factual learning from paraphrasing, whereas \citet{allen2024physics}, at batch size 96 with 512-token chunks, reports \textit{gains}. The two sit at opposite ends of the regime we characterize here.

\section{Prior Knowledge Matters}

\begin{table}[h]
\centering
\begin{tabular}{@{} l c c @{}}
\toprule
\textbf{Model} & \textbf{Base} & \textbf{CPT} \\
\midrule
OLMo-2-0425-1B & 0.4380 & 0.5454 \\
OLMo-2-1124-7B & 0.6859 & 0.7272 \\
\bottomrule
\end{tabular}
\caption{The prior-knowledge gap. MCQA accuracy on prerequisite topics for each document, before (\textit{Base}) and after (\textit{CPT}) continued pre-training on synthetic textbooks covering that foundational knowledge.}
\label{tab-prior-knowledge}
\end{table}

\textbf{Prior-Knowledge Gap.} We first establish that the model lacks some of the foundational knowledge $K$ presupposes. We generate MCQA pairs on prerequisite topics for each document and measure the base model's accuracy. As shown in Table~\ref{tab-prior-knowledge}, accuracy on this benchmark improves after continued pre-training on synthetic textbooks covering this material, confirming a gap that domain adaptation must contend with. Recent work has studied how such gaps shape learning~\citep{wang2025objectivefinetuningllmsprior, gekhman2024does, yang2024fine}, but the question remains open.

% \textbf{Distinct Benefits.} Adding either knowledge type on top of \textit{Para.\ 9} improves probe performance, but the two help different probe types (Figure~\ref{fig:prior_knowledge}). Contextual knowledge yields the larger gain on \textit{factual} probes in both domains, whereas prerequisite knowledge yields the larger gain on \textit{inference}. All conditions are token-matched as before.
\textbf{Complementary Benefits.} Table~\ref{tab:surrounding-knowledge} reports the peak improvement over each run's pretrained baseline. Relative to standard \textit{Para.\ 9}, providing either contextual or prerequisite knowledge substantially improves acquisition. Under the stricter token-matched comparison, adding surrounding knowledge does not consistently surpass \textit{Para.\ 9}. This is unsurprising; because contextual knowledge is largely tangential to the target material, allocating tokens to knowledge distant from the target knowledge should be less effective. Nevertheless, surrounding knowledge yields improvements comparable to adding paraphrases, a non-trivial benefit. Furthermore, the knowledge types exhibit distinct strengths: contextual knowledge drives larger factual gains, whereas prerequisite knowledge yields greater improvements in inference.

To ask whether these differences reflect surface overlap, we measure how often each probe's target span appears in the inserted texts (Table~\ref{tab:target-occurrence-insert}). Cited works include factual targets about twice as often as prerequisite textbooks, so contextual knowledge's relative factual advantage may partly reflect lexical presence. However, lexical overlap cannot explain the inference result: prerequisite knowledge contains the inference targets no more often than contextual knowledge, with lower bigram frequency, yet yields the larger inference improvement. This is consistent with prerequisite knowledge supplying foundations that support integration and inference.

\section{Ablations}

\textbf{Learning Rate.} Across our experiments, we use a fixed peak learning rate of 4e-5. Following \citet{parmar2024reuse}, however, continued-pre-training recipes should inherit the base model's pre-training learning rate, and the 13B and 32B models were pre-trained at higher rates (9e-5 and 6e-5) than the 7B model (3e-5)~\cite{olmo20242}. Holding it fixed at 4e-5 may have caused us to underestimate the model-size effect and the overall effects of auxiliary views and paraphrasing. Consistent with this interpretation, the analysis in Figure~\ref{fig:lr} shows that the advantages of auxiliary views widen as the learning rate increases.

\textbf{Order of Prior Knowledge.} We vary whether prerequisite data appears at the beginning, middle, or end of training. No placement is consistently best across metrics (Table~\ref{tab:prerequisite-order}), and the differences are small. We leave a fuller investigation of curriculum effects to future work.

\section{Discussion \& Conclusion}

\textbf{Auxiliary Views.} To our knowledge, we are the first to isolate the significance of auxiliary views. The premise is intuitive: data augmentation can help, and diversity is an established principle in pre-training. However, \textit{diversity remains an underspecified principle} that is difficult to operationalize when constructing data from scratch amid data scarcity. This motivates a clearer account of how variation improves learning and why.

Our study uncovers a phenomenon in which conceptual reformulations of a document, whether as a textbook or a blog, have an effect starkly distinct from mere paraphrasing: \textit{they substantially improve learning and induce a distinct layer-wise bias and compression in how knowledge is encoded}. More strikingly, they improve factual recall even though the model sees the original document, from which our probes are constructed, less frequently. This points to an underlying effect: \textit{a model's broader conceptual understanding directly facilitates its ability to memorize specific facts.}

Furthermore, the model's ability to encode auxiliary views effectively emerges only in our larger models and grows with model size. \textit{Larger models learn better by integrating diverse views more effectively}, encoding them with greater parameter efficiency by moving fewer weights and redistributing learning toward the middle and final layers. Together, these results help explain the effectiveness of pre-training corpora beyond simple scale and provide a clearer mechanism for the commonly invoked notion of ``good'' and ``diverse'' data.

Lastly, we notice that our result have implications for how much LLMs can self-improve. It is possible that LLMs can generate auxiliary views for itself and thus improve its knowledge on specialized, long-tailed topics, a possibility that warrants further investigation.

\textbf{Practical Takeaways.} We offer two recommendations for practitioners: (1) perform synthetic data augmentation with auxiliary views, and (2) train on missing prerequisite knowledge.

We find the first point especially relevant for scientific domains. Open scientific corpora are constantly evolving, producing research that lacks the auxiliary views which established knowledge are abundantly surrounded by in comparison. Our results show that synthesizing such views is helpful for acquiring knowledge effectively.

We hope these insights prove valuable to practitioners in domain adaptation, encourage greater consideration of how knowledge is represented, and provoke deeper discussion of the mechanisms of knowledge acquisition in LLMs.

\section*{Acknowledgements}

This work was supported in part by seed funding from the PSOM AI2D Center at Penn.

\section*{Limitations}

Our analysis is restricted to three domains, which may introduce corpus-level biases. While our main results regarding auxiliary views were consistent across each domain, this is not the case for our secondary results regarding contextual and prerequisite knowledge. We acknowledge that this limits the generalization of our findings.

Our findings may also have limited generalizability to pre-training. While our pre-training-faithful control uses the original optimizer state, learning-rate schedule, and data at a batch size of 1{,}024, it is a 100-step experiment at a later checkpoint rather than pre-training from scratch.

Furthermore, our investigation of generator strength may not generalize to low-resource, highly specialized domains, where even comprehending the source material may challenge the generator. In such settings, it remains unclear whether teacher-model strength is irrelevant.

Finally, our results are constrained by the scale of the models we study (up to 32B). Because model scale is a primary determinant of LLM capabilities, our findings may not extend to substantially larger models, though the patterns in our results suggest otherwise.

\bibliography{main}

\appendix
\section{Data}
\label{Appendix:preprocess}

\subsection{Dataset Construction}

\paragraph{Document collection.} The arXiv set was manually collected.  For medical documents, we use the NCBI E-utilities API to search PubMed Central for open-access case reports and download JATS XML.  For legal documents, we use the CourtListener REST API to search U.S. federal appellate opinions. Across domains, we filter candidate documents to obtain twelve target documents of suitable length, e.g., court opinions longer than 4000 tokens were rejected to avoid awkward chunking.

\paragraph{Pre-training corpus check.} We check all 36 source documents against the Infini-gram API. For each document, we query the title and ten randomly sampled body sentences against the OLMo-2 pre-training mixture index, \texttt{v4\_olmo-mix-1124\_llama}, and find zero matches.

\paragraph{Cleaning.}
For arXiv papers, we clean the raw LaTeX by removing comments, figures, tables, presentation-only commands, unresolved includes, page breaks, and bibliography/appendix material; expanding simple macros; extracting title, abstract, and main body; resolving revision commands; and repairing erroneous line breaks while preserving LaTeX environments.  Medical case reports are converted from JATS XML into sectioned plain text.  Legal opinions require additional PDF/OCR cleanup: we remove page headers, docket/page artifacts, decorative separators, extracted footnotes, and merge dangling sentences into paragraphs.

\paragraph{Contextual Knowledge.}
In our study, contextual views are cited works associated with the source document. For arXiv documents, we resolve each paper to an arXiv identifier, retrieve reference metadata using Semantic Scholar and OpenAlex, identify references with arXiv identifiers, download their source packages, and clean them with the same arXiv cleaning procedure. For legal documents, we use CourtListener citation links to collect cited judicial opinions and apply the same legal-opinion cleaning procedure. We do not construct contextual views for medical documents because case reports lack an analogous citation structure in our setup.

\subsection{Synthetic Data Generation}

\paragraph{Data statistics.}
Table~\ref{tab:dataset_artifact_inventory} summarizes the artifacts used in our experiments.  We generate 49 paraphrases per source document, yielding 1764 paraphrases in total.  For auxiliary views, $N$ counts the generated units obtained by splitting each view family into per-post blogs, per-question Stack Exchange Q\&A, and per-chapter textbooks.

\begin{table}[h]
\centering
\scriptsize
\setlength{\tabcolsep}{3pt}
\begin{tabular}{@{}lrrr@{}}
\toprule
\textbf{Material} & \textbf{$N$} & \textbf{Avg. Len. (tokens)} & \textbf{Total Tokens} \\
\midrule
\multicolumn{4}{@{}l}{\textit{Overall}} \\
Source Documents & 36 & 5858 & 210{,}888 \\
Paraphrases & 1764 & 6180 & 10{,}904{,}160 \\
Blogs & 225 & 1901 & 427{,}632 \\
Stack Exchange Q\&A & 453 & 1290 & 584{,}392 \\
Textbook Chapters & 282 & 2239 & 631{,}403 \\
Prerequisite Chapters & 582 & 4282 & 2{,}492{,}392 \\
Contextual Documents & 639 & 10416 & 6{,}655{,}888 \\
\midrule
\multicolumn{4}{@{}l}{\textit{Computer Science}} \\
Source Documents & 12 & 10550 & 126{,}598 \\
Paraphrases & 638 & 11002 & 7{,}019{,}384 \\
Blogs & 105 & 1920 & 201{,}648 \\
Stack Exchange Q\&A & 242 & 1460 & 353{,}321 \\
Textbook Chapters & 158 & 2872 & 453{,}833 \\
Prerequisite Chapters & 187 & 4555 & 851{,}724 \\
Contextual Documents & 506 & 10791 & 5{,}460{,}148 \\
\midrule
\multicolumn{4}{@{}l}{\textit{Medical}} \\
Source Documents & 12 & 3765 & 45{,}181 \\
Paraphrases & 588 & 3858 & 2{,}268{,}311 \\
Blogs & 72 & 1979 & 142{,}500 \\
Stack Exchange Q\&A & 117 & 1068 & 124{,}973 \\
Textbook Chapters & 61 & 1404 & 85{,}666 \\
Prerequisite Chapters & 220 & 4589 & 1{,}009{,}582 \\
Contextual Documents & 0 & -- & -- \\
\midrule
\multicolumn{4}{@{}l}{\textit{Legal}} \\
Source Documents & 12 & 3259 & 39{,}109 \\
Paraphrases & 589 & 3274 & 1{,}928{,}141 \\
Blogs & 48 & 1739 & 83{,}484 \\
Stack Exchange Q\&A & 94 & 1129 & 106{,}098 \\
Textbook Chapters & 63 & 1459 & 91{,}904 \\
Prerequisite Chapters & 175 & 3606 & 631{,}086 \\
Contextual Documents & 133 & 8991 & 1{,}195{,}740 \\
\bottomrule
\end{tabular}
\caption{Dataset statistics by text type and domain.}
\label{tab:dataset_artifact_inventory}
\end{table}

\paragraph{Paraphrases.}
Paraphrases are generated from cleaned source documents on a paragraph-level basis.  We preserve section headers verbatim and avoid paraphrasing LaTeX-only paragraphs. Domain-specific prompts are used for academic, legal, and medical text.  The academic prompt preserves LaTeX formatting, equations, proper nouns, titles, section headers, and technical terminology; the legal prompt preserves legal meaning, party names, citations, quoted language, dates, docket numbers, and legal terms of art; and the medical prompt preserves diagnoses, medications, doses, routes, timelines, lab values, imaging findings, procedures, outcomes, and clinical terminology. We use GPT-4.1 to generate paraphrases with a temperature of 1 and a top-p of 0.975.

\paragraph{Auxiliary views.}
For each document, we generate synthetic auxiliary views conditioned on the target document and intended to restate, explain, or pedagogically reorganize the same document-level content. The computer science setting generates textbook chapters for college students with a basic machine-learning background, Stack Exchange--style questions from a confused student followed by grounded answers, and technical blog posts for a broader technical audience. The medical setting adapts these formats to clinical education: case-based textbook sections for medical students or residents, clinical teaching Q\&A, and clinical blog posts. The legal setting uses casebook/treatise-style chapters for law students, Law Stack Exchange--style Q\&A, and analytic legal commentary. For each view family, we first generate an outline, list of questions, or list of blog ideas and then generate the individual chapters, answers, or posts from that plan. We use GPT-5 to generate the outlines and GPT-5-mini to generate the contents of the chapters, questions, or posts.

\paragraph{Prerequisite views.}
Prerequisite views are generated separately from auxiliary views.  For arXiv papers, we use a curriculum-design prompt to create prerequisite textbook chapters that teach the foundations needed to understand the paper while excluding the paper's own novel ideas.  For medical case reports, the prompt asks for general medical background--for example relevant anatomy, pathophysiology, pharmacology, diagnostic interpretation, differential diagnosis, and standard management--while explicitly forbidding patient-specific chronology, workup, treatment course, outcome, or novel observations from the source case.  For legal opinions, the prompt similarly asks for doctrinal and procedural background while excluding the source case's parties, facts, outcome, holding, and novel reasoning. When useful for legal background, we additionally include landmark or doctrinally foundational cited opinions as context for the generated prerequisite chapters. We use GPT-5 to generate the outlines and GPT-5-mini to generate the contents of the chapters.

\FloatBarrier

\section{Probe Generation}
\begin{figure*}[t]
    \centering
    \includegraphics[width=0.95\textwidth]{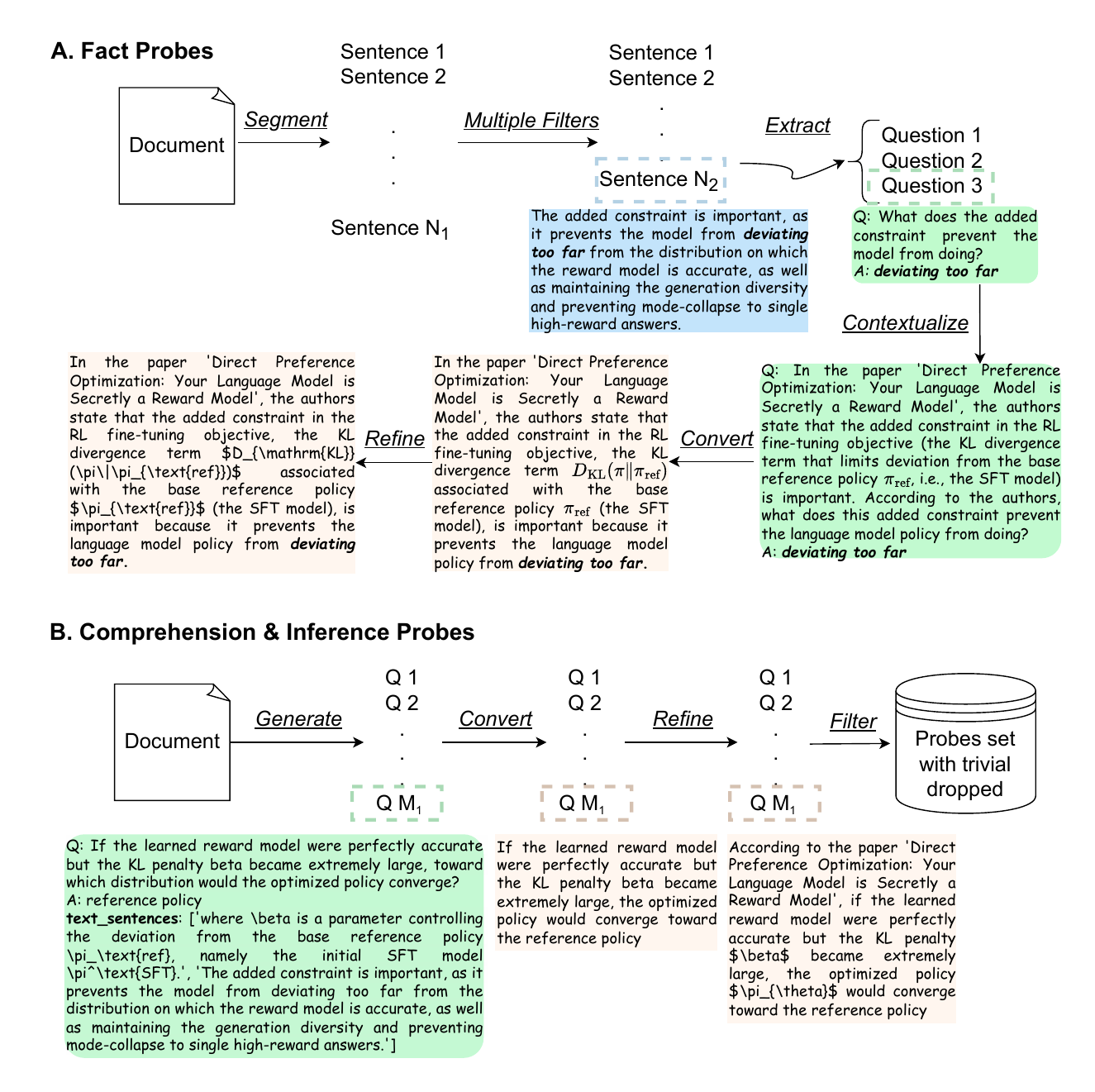}
\caption{Overview of the probe-generation pipeline for arXiv documents. \textbf{(A) Factual probe generation}: We generate probes sentence by sentence to keep their ground truth close to the source document. \underline{Preprocessing}: We first remove sentences containing minimal knowledge (to prevent noise from structural comments, e.g., ``Let us first discuss the following results''). We then use heuristics to remove sentences likely to yield low-quality probes, such as those that are too short (e.g., ``the sweep has 22 runs'') or contain excessive \LaTeX{} code with no valid English extraction targets. \underline{Question extraction}: From each remaining sentence, we extract 1--3 questions that capture its knowledge. \underline{Contextualization}: Questions are made clear and self-contained while preserving the knowledge being tested. \underline{Cloze conversion}: Questions are converted into cloze statements with answers at the end. \underline{Refinement}: We ensure that all mathematical content is written in \LaTeX{} and verify the preceding steps. \textbf{(B) Compositional probe generation}: Because not all atomic facts warrant a compositional probe, we employ a two-level approach. First, we divide the paper into sections and prompt an LLM to extract compositional questions. We define compositionality as either (1) \textbf{inference}, which reasons over supporting text to reach a new insight, or (2) \textbf{synthesis}, which combines several facts into a synthesized statement. We perform this process section by section for granularity, then repeat it with the entire paper to obtain more holistic questions. \underline{Cloze conversion}: As in the factual pipeline, questions are converted into cloze statements with answers at the end. \underline{Refinement}: Statements are formatted in \LaTeX{} where needed and made self-contained by referencing the paper (e.g., ``according to the paper''). \underline{Filtering}: Finally, we ask the LLM to identify questions that are too simple, imprecise, or confusing; this step removes 15\% of probes on average.}
    \label{figA1}
\end{figure*}

\FloatBarrier
Our probe construction pipeline is adapted and refined for each domain (arXiv, legal, and medical). We present the pipeline for arXiv documents in Figure~\ref{figA1}; the full pipeline for all domains is available in our released code.

\FloatBarrier
\clearpage
\twocolumn[{
\section{Additional Plots}
%
% \begin{figure*}[h]
%     \centering
\begin{center}
    \includegraphics[width=\textwidth]{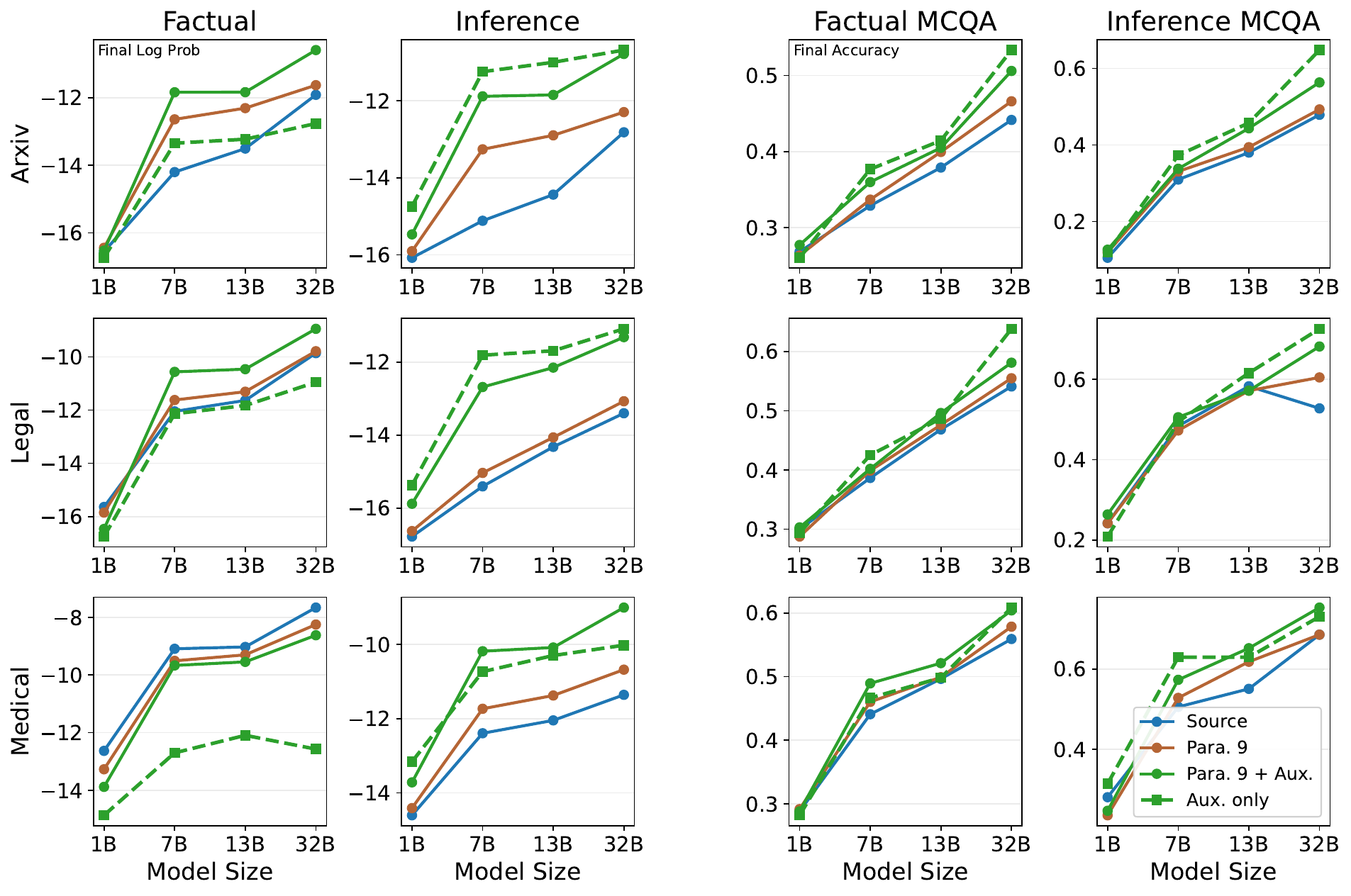}
    \captionof{figure}{Knowledge acquisition by domain under a fixed token budget. Solid green circles denote \textit{Para. 9 + Aux.} training, and dashed green squares denote \textit{Aux. only} training. Auxiliary views consistently improve factual recall and inference across domains, except that \textit{Aux. only} training has lower factual cloze log probability. \textit{Para. 9 + Aux.} training also performs worse on medical factual probes.}
    \label{views_by_domain}

    \vspace{1em}

    \includegraphics[width=0.75\textwidth]{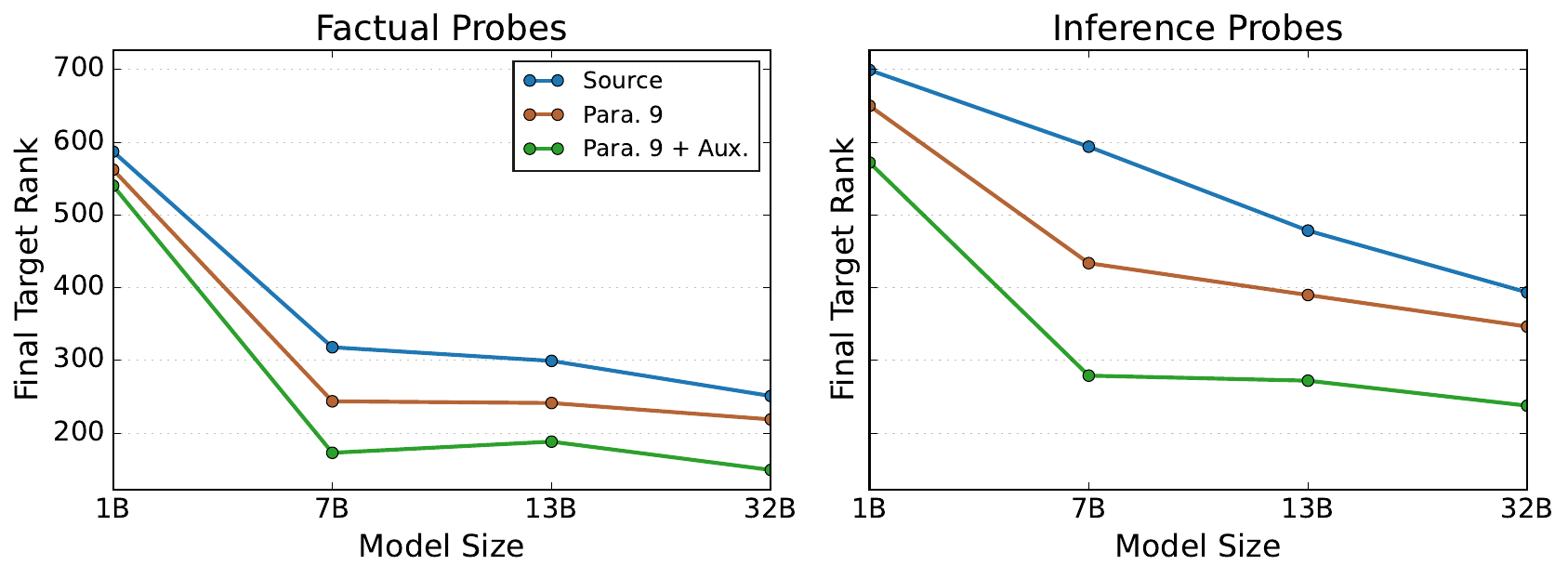}
    \captionof{figure}{Under a fixed token budget, auxiliary views improve both factual recall and inference, and their effect size increases with model size. Lower ranks are better.}
    \label{views_target_rank}

\end{center}
}]

\clearpage

\begin{figure*}[!t]
    \centering
    \includegraphics[width=\textwidth]{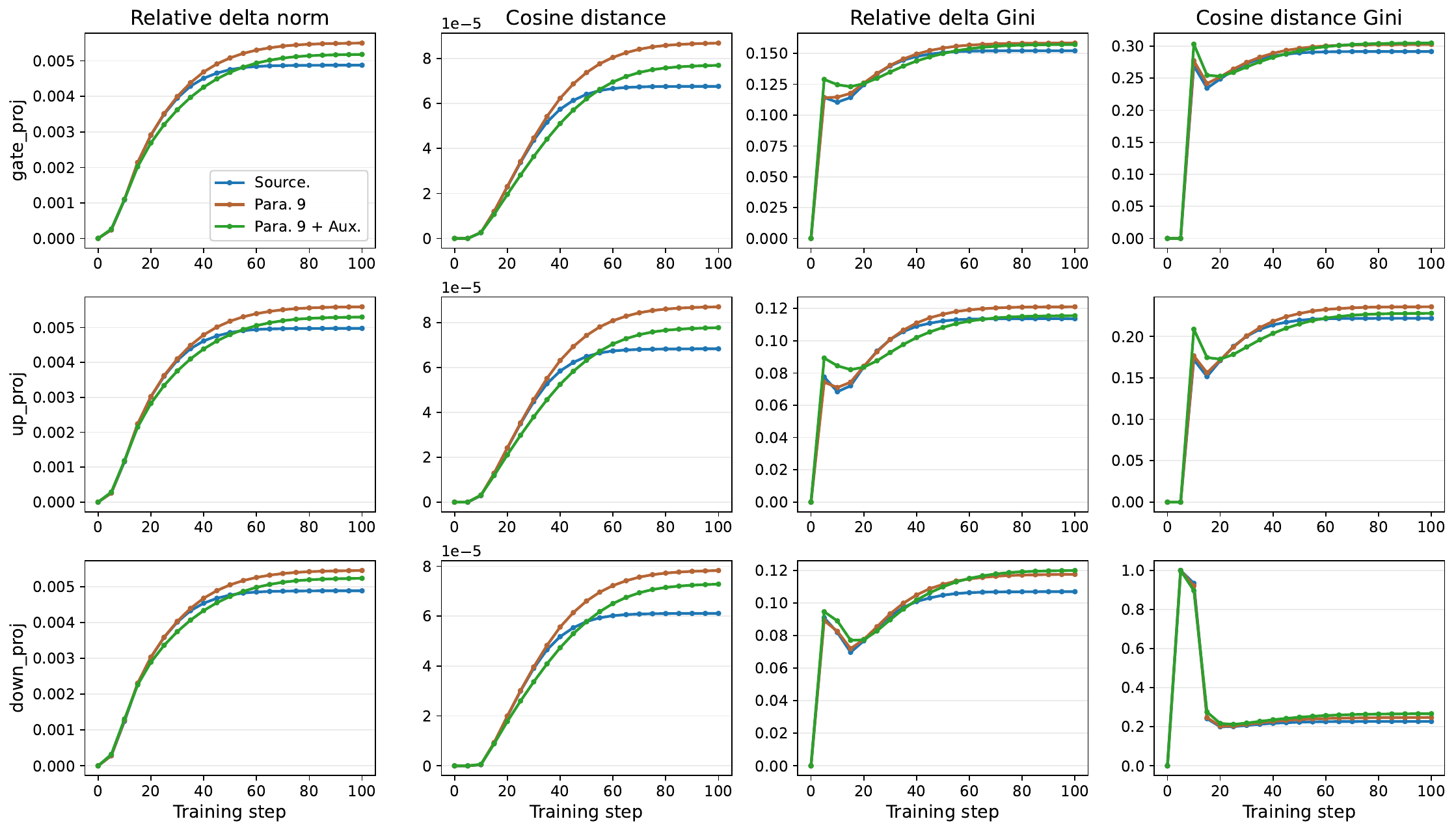}
\caption{FFN parameter change over training (OLMo-2-7B). Paraphrasing induces the largest parameter movement, while auxiliary views reduce this magnitude toward source-only training. Both the concentration of change (Gini) and downstream performance converge by step \(\sim\)50 (performance not shown), whereas magnitude (relative delta norm and cosine distance) continues to grow afterward; parameters keep drifting after learning has effectively converged.}
    \label{fig:mech_steps}
\end{figure*}

\begin{figure*}[!t]
    \centering
    \includegraphics[width=\textwidth]{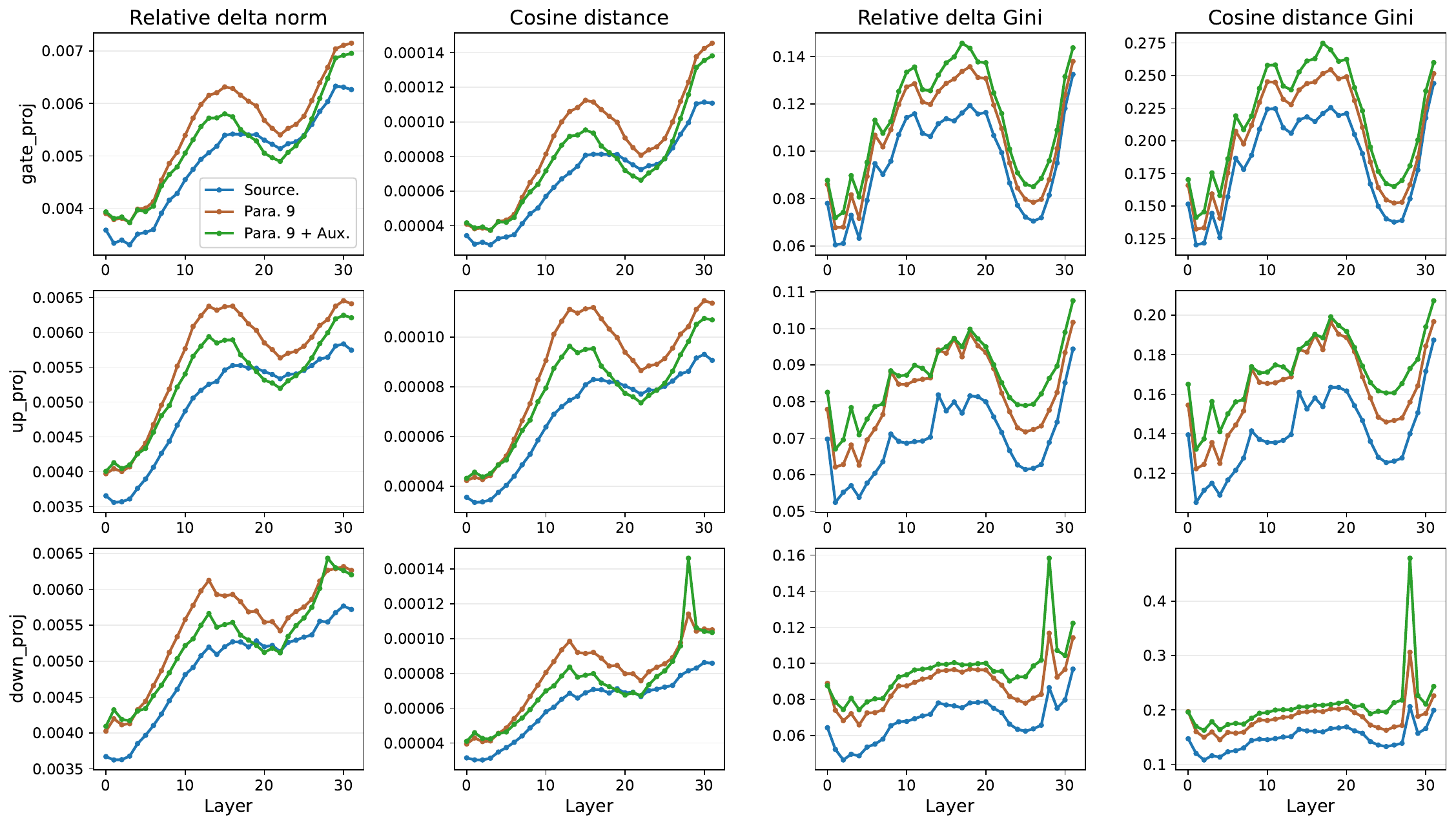}
\caption{Per-layer FFN parameter change from the base model over training (OLMo-2-7B). \textit{Magnitude} (relative delta norm and cosine distance) and \textit{concentration} (Gini) are shown for the gate, up, and down projections. All conditions concentrate change in the middle and final layers with a dip in the upper-middle layers; auxiliary views increase change in the middle and final layers but reduce it relative to source in the \(\sim\)16--24 band.}
    \label{fig:mech_layer}
\end{figure*}

\begin{figure*}[!t]
    \centering
    \includegraphics[width=0.85\textwidth]{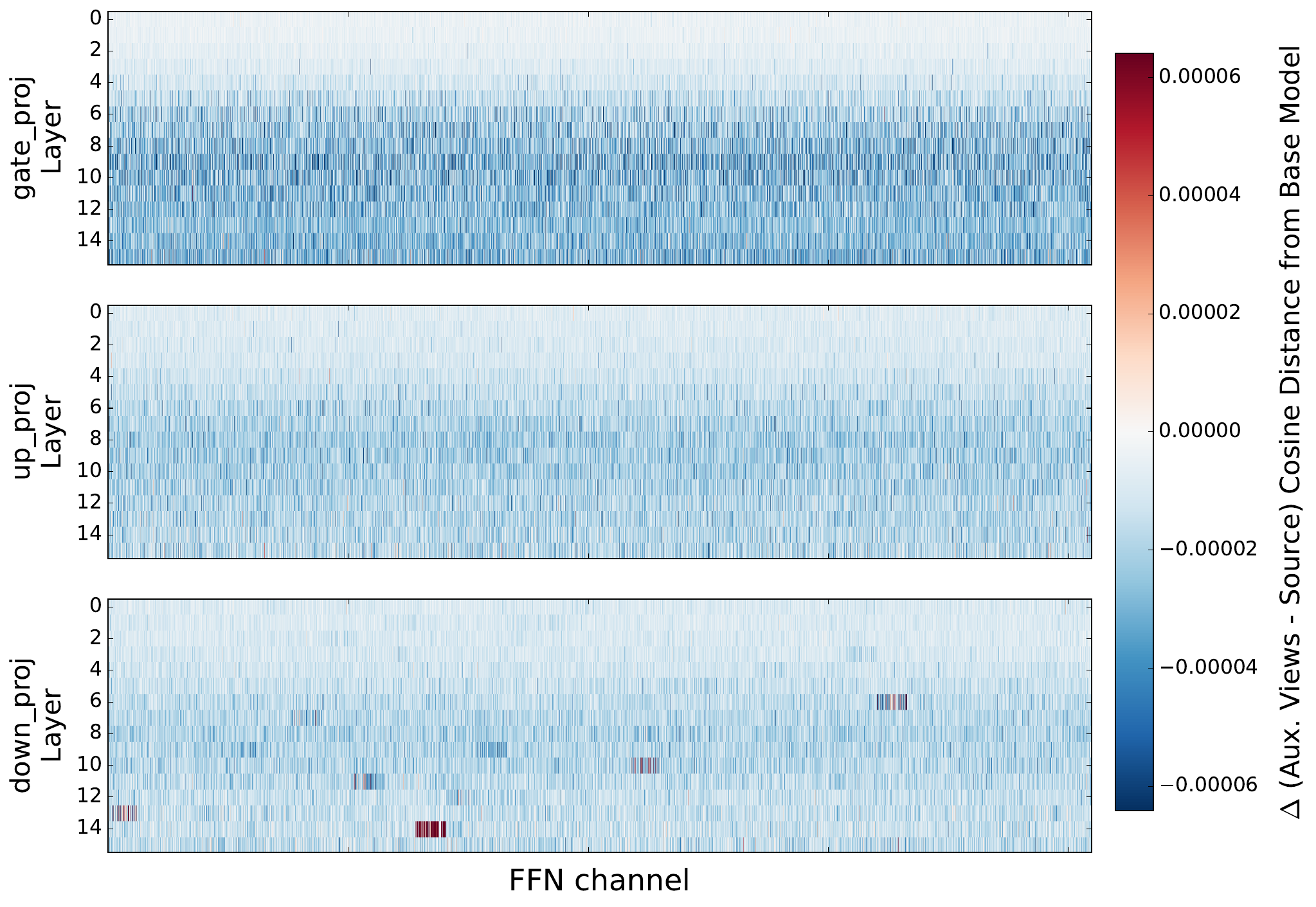}
\caption{The same per-channel difference in cosine distance from the base model (\textit{Para. 9 + Aux.}\ minus \textit{Source}) as Figure~\ref{fig:mech_heatmap}, but for the 1B model. Unlike the structured pattern at 7B, the difference is either blue (negative) or white, indicating that auxiliary views change parameters slightly \textit{less} than source nearly everywhere, with no organized layer-wise band. The only exceptions are a few isolated channels in \texttt{down\_proj} at the deepest layers. }
\label{fig:mech_heatmap_1b}
\end{figure*}

\begin{figure*}[!t]
    \centering
    \includegraphics[width=\textwidth]{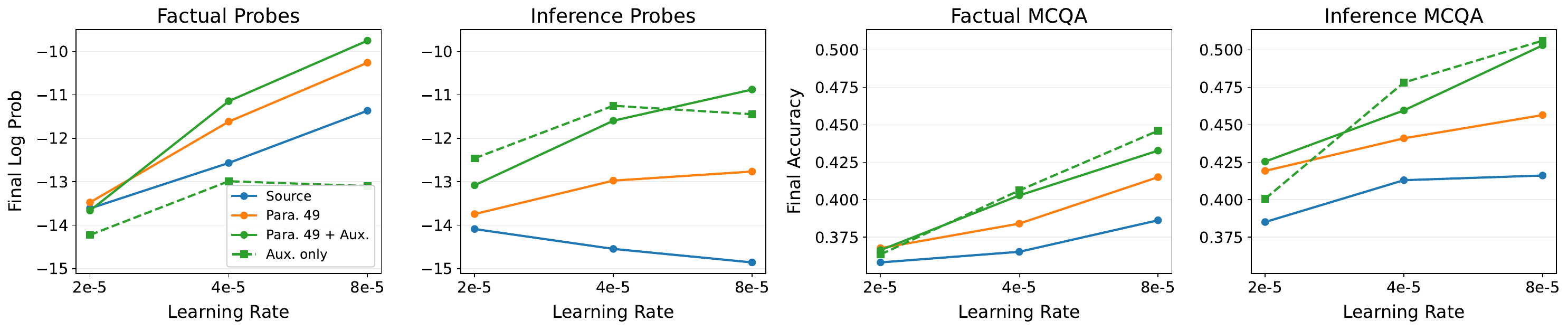}
\caption{Effect of peak learning rate (2e-5, 4e-5, and 8e-5) on knowledge acquisition for the 7B model. We report final log probability on factual and inference probes (left two panels) and final accuracy on factual and inference MCQA (right two panels), comparing source documents alone (\textit{Source}), 49 paraphrases (\textit{Para.\ 49}), and 49 paraphrases plus auxiliary views (\textit{Para. 49 + Aux.}). Across all metrics, the advantages of auxiliary views and paraphrases grow with learning rate.}
\label{fig:lr}
\end{figure*}

% \begin{figure*}[!t]
%     \centering
%     \includegraphics[width=0.8\textwidth]{figures/prior_knowledge_match_histogram_vs_para_7B.pdf}
% \caption{Gain in \(\Delta\) log prob.\ over \textit{Para.\ 9} from adding prerequisite versus contextual knowledge, by domain and probe type. Medical is omitted (no citation structure). Contextual knowledge helps factual probes more; prerequisite knowledge helps inference probes more.}
% \label{fig:prior_knowledge}
% \end{figure*}

% \begin{figure*}[!t]
%     \centering
%     \includegraphics[width=0.9\textwidth]{figures/prior_knowledge_position_logprob_histogram_delta_all_domains_7B.pdf}
% \caption{Effect of prerequisite-concept ordering on knowledge acquisition. We compare placing prerequisite data at the \textit{Front}, \textit{Middle}, or \textit{End} of training, reporting the change in log-probability ($\Delta$ Log Prob) on factual (left) and inference (right) probes. Ordering has only a minor effect, with \textit{Front} placement performing slightly worse than \textit{Middle} or \textit{End}.}
% \label{fig:order}
% \end{figure*}

\FloatBarrier
\section{Additional Tables}

\begin{table}[!htbp]
\centering
\scriptsize
\setlength{\tabcolsep}{3pt}
\resizebox{\columnwidth}{!}{
\begin{tabular}{@{}lcccc@{}}
\toprule
\textbf{Auxiliary views} & \textbf{Factual} & \textbf{Factual} & \textbf{Inference} & \textbf{Inference} \\
 & \textbf{log prob.} & \textbf{MCQA} & \textbf{log prob.} & \textbf{MCQA} \\
\midrule
Textbooks & -11.358 & 0.394 & -12.218 & \textbf{0.457} \\
Stack Exchange & -11.642 & 0.398 & -12.216 & 0.450 \\
Blogs & -11.570 & 0.388 & -12.189 & 0.447 \\
Mixed & \textbf{-11.087} & \textbf{0.398} & \textbf{-11.665} & 0.450 \\
\bottomrule
\end{tabular}
}
\caption{\textbf{Auxiliary-view families.} Metrics for OLMo-2 7B after training on token-matched textbooks, Stack Exchange--style Q\&A, blogs, or their mixture. The families perform similarly; mixing them performs best on both log-probability metrics and factual MCQA, while textbooks perform best on inference MCQA. Higher is better for every metric.}
\label{tab:view-family}
\end{table}

\begin{table}[!htbp]
\centering
\scriptsize
\setlength{\tabcolsep}{3pt}
\resizebox{\columnwidth}{!}{
\begin{tabular}{@{}llcc@{}}
\toprule
\textbf{Domain} & \textbf{Condition} & \textbf{Peak factual $\Delta$ LP} & \textbf{Peak inference $\Delta$ LP} \\
\midrule
\multirow{4}{*}{arXiv}
 & Para.\ 9 (standard) & 4.344 & 2.575 \\
 & Para.\ 9 (token-matched) & \textbf{7.117} & \textbf{4.723} \\
 & $+$ Prerequisite & 6.690 & 4.629 \\
 & $+$ Contextual & 6.893 & 4.574 \\
\midrule
\multirow{4}{*}{Legal}
 & Para.\ 9 (standard) & 3.817 & 0.719 \\
 & Para.\ 9 (token-matched) & \textbf{8.222} & 2.049 \\
 & $+$ Prerequisite & 6.663 & \textbf{2.331} \\
 & $+$ Contextual & 6.768 & 2.184 \\
\midrule
\multirow{4}{*}{All}
 & Para.\ 9 (standard) & 4.231 & 2.028 \\
 & Para.\ 9 (token-matched) & \textbf{7.349} & 3.950 \\
 & $+$ Prerequisite & 6.684 & \textbf{3.967} \\
 & $+$ Contextual & 6.866 & 3.881 \\
\bottomrule
\end{tabular}
}
\caption{\textbf{Contextual and prerequisite knowledge.} Peak improvement in log probability (LP) from each run's pretrained baseline. The added-knowledge conditions and token-matched \textit{Para.\ 9} use the same inserted-token budget; standard \textit{Para.\ 9} is included as a non-token-matched reference. Adding surrounding knowledge yields improvements comparable to adding paraphrases. Higher $\Delta$ LP is better. Medical is omitted because case reports lack an analogous citation structure.}
\label{tab:surrounding-knowledge}
\end{table}

\begin{table}[!htbp]
\centering
\scriptsize
\setlength{\tabcolsep}{3pt}
\resizebox{\columnwidth}{!}{
\begin{tabular}{@{}lcccc@{}}
\toprule
\textbf{Placement} & \textbf{Factual} & \textbf{Factual} & \textbf{Inference} & \textbf{Inference} \\
 & \textbf{$\Delta$ LP} & \textbf{$\Delta$ MCQA} & \textbf{$\Delta$ LP} & \textbf{$\Delta$ MCQA} \\
\midrule
Front & 6.320 & 0.022 & 3.639 & \textbf{0.068} \\
Middle & 6.591 & \textbf{0.026} & \textbf{3.847} & 0.053 \\
End & \textbf{6.680} & 0.025 & 3.706 & 0.053 \\
\bottomrule
\end{tabular}
}
\caption{\textbf{Prerequisite-knowledge ordering.} Peak improvement from the pretrained baseline when prerequisite data appears at the front, middle, or end of training. Higher is better for every metric, including $\Delta$ log probability (LP). No placement is consistently best.}
\label{tab:prerequisite-order}
\end{table}

\begin{table}[h]
\centering
\small
\setlength{\tabcolsep}{4pt}
\renewcommand{\arraystretch}{0.9}
\caption{Frequency and coverage of our probe targets across documents. \textit{Freq.} = occurrences per 1k tokens; \textit{Cover.} = fraction of targets present at least once across the documents. \textit{Para.\ 49} averages \textit{Freq.} over 49 paraphrased documents, while coverage records presence in any paraphrase.}
\label{tab:target-occurrence-all}
\begin{tabular}{llrrrr}
\toprule
Probes & Corpus & \multicolumn{2}{c}{Full target} & \multicolumn{2}{c}{Bigram} \\
\cmidrule(lr){3-4}\cmidrule(lr){5-6}
 & & Freq. & Cover. & Freq. & Cover. \\
\midrule
\multirow{3}{*}{Factual} & Source & 0.391 & 0.70 & 0.829 & 0.92 \\
 & Para. 49 & 0.222 & 0.60 & 0.681 & 0.91 \\
 & Aux. & 0.112 & 0.33 & 0.437 & 0.74 \\
\cmidrule(lr){1-6}
\multirow{3}{*}{Inference} & Source & 1.225 & 0.51 & 0.703 & 0.71 \\
 & Para. 49 & 1.111 & 0.56 & 0.605 & 0.78 \\
 & Aux. & 1.070 & 0.51 & 0.628 & 0.77 \\
\bottomrule
\end{tabular}
\end{table}

\begin{table}[!htbp]
  \centering
  \small
  \setlength{\tabcolsep}{4pt}
  \renewcommand{\arraystretch}{0.9}
  \caption{Frequency and coverage of our probe targets across contextual and prerequisite knowledge. \textit{Freq.} = occurrences per 1k (words for the full target, OLMo
  tokens for bigrams); \textit{Cover.}\ = fraction of targets present at least once (full
  target) or the mean fraction of a target's bigrams present (bigram). \textit{Prerequisites} = generated textbook chapters; \textit{Cited Works} = cited papers or legal opinions.}
  \label{tab:target-occurrence-insert}
  \begin{tabular}{llrrrr}
  \toprule
  Probes & Corpus & \multicolumn{2}{c}{Full target} & \multicolumn{2}{c}{Bigram} \\
  \cmidrule(lr){3-4}\cmidrule(lr){5-6}
   & & Freq. & Cover. & Freq. & Cover. \\
  \midrule
  \multirow{2}{*}{Factual} & Prerequisites  & 0.031 & 0.12 & 0.129 & 0.54 \\
   & Cited Works & 0.039 & 0.24 & 0.141 & 0.72 \\
  \cmidrule(lr){1-6}
  \multirow{2}{*}{Inference} & Prerequisites  & 0.827 & 0.21 & 0.154 & 0.47 \\
   & Cited Works & 0.838 & 0.25 & 0.095 & 0.54 \\
  \bottomrule
  \end{tabular}
  \end{table}

\begin{table*}[!t]
\centering
\footnotesize
\setlength{\tabcolsep}{5pt}
\resizebox{0.98\textwidth}{!}{
\begin{tabular}{@{}lcccc@{}}
\toprule
\textbf{Generator} & \textbf{Generator factual} & \textbf{Params} & \textbf{Words} & \textbf{Downstream factual} \\
 & \textbf{MCQA acc.} & \textbf{(B)} & \textbf{(M)} & \textbf{MCQA acc.} \\
\midrule
None (pretrained OLMo-2 7B) & -- & -- & -- & 0.337 \\
Para.\ 9 (token-matched baseline) & -- & -- & -- & 0.377 \\
\midrule
gpt-5-mini (original; mixed) & -- & -- & 1.08 & 0.398 \\
gpt-5-mini (low reasoning) & 0.657 & -- & 1.04 & 0.403 \\
gpt-5-mini (high reasoning) & 0.673 & -- & 1.10 & 0.399 \\
gpt-5.4-mini (low reasoning) & 0.718 & -- & 0.84 & 0.397 \\
gpt-5.4-mini (high reasoning) & 0.749 & -- & 0.72 & 0.396 \\
gpt-oss-20B (low reasoning) & 0.539 & 20.9 & 0.72 & \textbf{0.405} \\
gpt-oss-120B (low reasoning) & 0.599 & 116.8 & 0.79 & 0.394 \\
Gemma-4 12B IT & 0.565 & 12 & 0.36 & 0.392 \\
Gemma-4 31B IT & 0.662 & 31 & 0.35 & 0.386 \\
GLM-5 (high reasoning) & 0.728 & 744 & 0.66 & 0.385 \\
GLM-5.2 (high reasoning) & 0.758 & 744 & 0.90 & 0.400 \\
\bottomrule
\end{tabular}
}
\caption{\textbf{Auxiliary-view generation.} Peak factual MCQA accuracy after training OLMo-2 7B on views from different generators under the same schedule and token budget. Generator accuracy measures five-shot prior domain knowledge; Words (M) counts generated blogs, Stack Exchange posts, and textbooks. Downstream accuracy is uncorrelated with generator size among open-weight generators (Pearson $r=-0.11$, $n=6$) or generator accuracy ($r=-0.18$, $n=10$), but correlates moderately with view-text volume ($r=+0.68$, $n=11$, $p=0.020$).}
\label{tab:generator-ablation}
\end{table*}

\FloatBarrier
\section{Hyperparameters for Replication and Details for Reproduction}\label{append:hyperp}
For training, we use the TRL library. Unless otherwise noted, our experiments use the following defaults:
learning rate $4\times10^{-5}$, context size 4096, batch size 256, weight decay 0.1,
cosine decay scheduler with 0.1 warm-up ratio and minimum learning rate ratio of 0.1, seed 42, max gradient norm of 1, and the AdamW optimizer
($\beta_1 = 0.9$, $\beta_2 = 0.999$, $\epsilon = 10^{-8}$) with BF16 training.

\textbf{Pre-training-faithful continuation.} For the experiment in Table~\ref{tab:proper-pretraining}, we resume OLMo-2 7B from checkpoint step 925{,}000 using the OLMo framework. We restore the checkpoint's optimizer state and continue with the original pre-training data stream and learning-rate schedule, a global batch size of 1{,}024, and a sequence length of 4{,}096. We inject domain data for 100 optimizer steps using the same conditions as in the main experiments.

\section{Prompts for Probe Construction (arXiv)}\label{Appendix:Probes}

We provide some of the prompts used in our pipeline to construct factual and inference probes for the arXiv documents. The complete set of prompts, adapted for each domain (legal and medical), is available in our code.%\syang{knowledge probe, inference probe? please revise the text here and the prompts descriptions}

\textbf{1. Prompt to extract atomic facts from a paragraph:}
\begin{lstlisting}[style=prompt]
You will be given two inputs, a section of an academic paper for context and a single sentence drawn from that section. Papers often interweave various pieces of knowledge together in academic writing. While each sentence is interwoven with others, there is atomic knowledge that can be extracted from a particular sentence. Write questions that tests for this atomic knowledge. Specifically, your task is to extract questions from the provided sentence with clear answers, each 1 to 4 words long. 
   
Extract 1-3 questions from the sentence. 
  
### Detailed Instructions
Consider these instructions as you extract each question:
- The question should be natural and meaningful, in which the answer is considered a main fact presented by the sentence.
- The answer should be non-trivial and non-obvious. It should not be deducible from the sentence itself.
- The answer to the question should be a meaningful, coherent phrase, 1-4 words long, taken from the sentence. Simplify the answer by stripping determiners such as "some" or "a" or "an" or "the" from the answer. Feel free to adjust the answer to fit the question, but the meaning should be the same.
- The answer must *NOT* involve any *special characters* or *mathematical notation*. Again, any question with an answer that contains mathematical notation should not be used.
- The question should have a a clear, single answer and *NOT* multiple valid answers. 
- Each question should be written separately and independently of the other questions, so don't reference other questions in the same question.

### Demonstration 1
Context: "\\title{Direct Preference Optimization: Your Language Model is Secretly a Reward Model}\n\\subsection{Can DPO scale to real preference datasets?}\nNext, we evaluate fine-tuning performance of DPO on summarization and single-turn dialogue. For summarization, automatic evaluation metrics such as ROUGE can be poorly correlated with human preferences~\citep{stiennon2022learning}, and prior work has found that fine-tuning LMs using PPO on human preferences to provide more effective summaries. We evaluate different methods by sampling completions on the test split of TL;DR summarization dataset, and computing the average win rate against reference completions in the test set."

Sentence: "We evaluate different methods by sampling completions on the test split of TL;DR summarization dataset, and computing the average win rate against reference completions in the test set."

Questions:
- "The authors evaluate DPO's fine-tuning performance against other methods on summarization by sampling completions on the test split of what dataset?", Answer: "TL;DR summarization"
- "The fine-tuning performance of DPO and other methods on summarization are evaluated by sampling completions on the test split of the TL;DR summarization dataset and computing the average win rate against what?", Answer: "reference completions"

### Demonstration 2
Context: "\title{Direct Preference Optimization: Your Language Model is Secretly a Reward Model}\nWhile large-scale unsupervised language models (LMs) learn broad world knowledge and some reasoning skills, achieving precise control of their behavior is difficult due to the completely unsupervised nature of their training. Existing methods for gaining such steerability collect human labels of the relative quality of model generations and fine-tune the unsupervised LM to align with these preferences, often with reinforcement learning from human feedback (RLHF)."

Sentence: "Existing methods for gaining such steerability collect human labels of the relative quality of model generations and fine-tune the unsupervised LM to align with these preferences, often with reinforcement learning from human feedback (RLHF)."

Questions:
- "What do existing methods collect to steer unsupervised language models, ?", Answer: "human labels"
- "Existing methods for steering unsupervised language models collect human labels of the quality of what?", Answer: "relative quality of model generations"
- "Existing methods align unsupervised language models by fine-tuning on what?", Answer: "human preferences"
- "Existing methods for steering unsupervised language models via fine-tuning on human preferences often use what?", Answer: "RLHF"
\end{lstlisting}

\textbf{2. Prompt to extract atomic facts from a paragraph with context:}
\begin{lstlisting}[style=prompt]
You will be given two inputs, a section of an academic paper for context, a single sentence drawn from that section, and a question extracted from the sentence as well as its corresponding answer. Your task is to then turn the question into a self-contained, precise question. Approach this task step-by-step as outlined below.

While you should use your expertise on this domain to handle and understand these texts, all information written into the questions and answers *MUST* originate from the provided context or sentence. Do not add, infer, or correct information using your internal knowledge. Every detail should be traceable back to the source text. As you write and rewrite the questions, also make sure to accurately represent the knowledge in the original sentence without distortion. Strive to use phrasing as close as possible to the original text, but prioritize clarity and self-containment. Lastly, the questions should be written well and clearly so that they are easy to read.
    
### Instructions
The overall goal of this task is to make the questions clear by incorporating the relevant context. This ensures the question is unambiguous and doesn't require looking back to the source material.

For each question:
1.  Rewrite the question so that it starts with one of the following templates. 
    - "In the paper '{title}', ..."
    - "According to the paper '{title}',..."
    - "In the paper '{title}', the authors remark that..."
    - "In the paper '{title}', the authors state that..."
    - "According to the paper '{title}', prior work has..."
    - "In the theoretical analysis of the paper "{title}"..."
    - "In the paper '{title}', the results suggest that..."
    This is a non-exhaustive list of templates, and you should use your own judgement to choose the most appropriate template or modify the template to fit the sentence.
2.  Add sufficient context. Specifically, use the *provided context* to supply whatever information is needed to make the question self-contained and unambiguous. For instance, "Do humans and GPT4 agree often with each other?" should be clarified into "In the paper '...', did humans and GPT4 often agree or disagree with each other during the evaluation of DPO?" if this notion was in the context of evaluating DPO in an academic paper. The goal is to ensure someone reading just the question would understand exactly what is being asked without needing additional context.
3.  Clarify pronouns and referential terms. Check the sentence for pronouns (it, this, that, these, those) or demonstrative phrases (this equation, that method, these results) that refer to entities not explicitly defined within the sentence itself. Search the surrounding context to identify what these terms reference, then incorporate that clarifying information into the question to make it self-contained.
4.  Clarify Context-Dependent Terms. Named entities (e.g., theorems, equations, proper nouns) do not need clarification. However, if there are unnamed or context-specific terms (e.g., $f$, "the model", "the loss"), clarify their full context. For instance, "the gradient" might refer to the general concept of a gradient or to the gradient of a specific function mentioned earlier in the context.
5.  Disambiguate experiments. There are often numerous experiments in a paper, and so supply enough experimental context so that the question is about which experiment the question is asking about. 
6.  Handle acronyms. If the answer is an acronym and the acronym appears frequently in the context, feel free to leave it as an acronym without defining it.
7.  Do not leak the answer. Please make sure that *the answer is not revealed* in the question. The answer should never appear in the question.
8.  Maintain the essence of the original question during all of this.
9.  Do not change the answer. Minor grammatical adjustments to the answer are allowed only if necessary to fit the restructured question (e.g., adjusting verb tense, determiners like "the").
10. Avoid quoting the source sentence directly in the question.
11. Refine Question. The rewritten question can be broken up into multiple sentences if the question becomes verbose. Make sure the question is written clearly and grammatically correct. Do not put any of the context in parenthesis or followed after an "i.e.".

Think carefully and critically through this task, following the step-by-step instructions outlined above. Then, provide the final output, listing each question and its corresponding answer.
\end{lstlisting}

\textbf{3. Prompt to generate inference questions from the text:}
\begin{lstlisting}[style=prompt]
You have been given a section of an academic text. Your tasks is to test the reader's understanding of the text. However, you should not test anything that can be recalled from reading a single sentence. Create questions that integrate, connect, and synthesize information across several sentences and aim at measuring a deeper understanding. Your question must not be obvious from a single sentence already in the paper, and truly require several sentences to synthesize the answer. Lastly, the answer to the question must be a coherent phrase, from 1 to 5 words long.

For each question, show me the sentences in the text that you're pulling from to answer the question. The question should be non-obvious from these sentences and require composing information from all of them to answer the question.

Provide the output in JSON format, as a dictionary with a single key "qa_items" which is a list of dictionaries with the following keys:
- "question": (string) 
- "answer": (string)
- "text_quotes": list of sentences from the text that you're pulling from to answer the question.
\end{lstlisting}

\section{Prompts for Synthetic Data Generation (arXiv)}\label{Appendix:Prompts} %(paragraph, prior knowledge, explanations)}
% We paraphrase each arxiv paper by asking the LLM with temperature=1 and top\_p=0.95.

\textbf{1. Prompt to synthesize Stack Exchange--style question--answer pairs:}

Question generation:
\begin{lstlisting}[style=prompt]
You are a confused student reading this research paper. You are struggling with specific concepts, details, and connections in this paper. Generate a list of several Stack Exchange style questions that you would ask to clarify your understanding.

Your questions should:
- Vary in levels of understanding, from misled to profound.
- Vary in complexity, from simple to deep.
- Vary in type, from conceptual to detail-specific.
- Focus on clarifying the concepts and details of the paper. Do not ask tangential questions.

As you generate the questions, please make sure to consider the following:
- Make sure the questions are self-contained and unambiguous
- Please write any mathematical notation in LaTeX only e.g. "$x^2$" or "$\pi$". Do not use unicode mathematical characters e.g. "pi".

For each question, provide:
- A `title` in Stack Exchange question format
- The `question_body` with context and what specifically you're confused about

## Example Question

"How can Transformers handle arbitrary length input?

The transformer, introduced in the paper Attention Is All You Need, is a popular new neural network architecture that is commonly viewed as an alternative to recurrent neural networks, like LSTMs and GRUs.

However, having gone through the paper, as well as several online explanations, I still have trouble wrapping my head around how they work."

### Output Format
Provide the output as a JSON object with a single key "questions", which is a list of question dictionaries.
Example:
{
  "questions": [
    {
      "title": "Why does the partition function cancel out in DPO derivation?",
      "question_body": "I'm reading the DPO paper and I understand that they start with the KL-regularized objective, but I'm confused about how the partition function Z(x) cancels out when they move to pairwise preferences. Can someone explain this step intuitively?",
    }
  ]
}
\end{lstlisting}

Answer generation:
\begin{lstlisting}[style=prompt]
A graduate student has asked a question about a research paper. Provide a clear, detailed Stack Exchange style answer that:

- Thoroughly addresses their question 
- Don't make it too lengthy; it should be concise and to the point like a Stack Exchange answer
- Write in prose rather than structured bullet points in one cohesive answer
- Provides intuitive explanations alongside technical details
- Connects to broader concepts when relevant
- Is educational and accessible

Please write any mathematical notation in LaTeX only e.g. "$x^2$" or "$\pi$". Do not use unicode mathematical characters e.g. "pi". Also, please make sure that your answer is grounded in the paper; do not provide any information that is inconsistent with the paper.

Again, please write all math in LaTeX.

Format your response as a comprehensive Stack Exchange answer.

### Example

Question:
"I know that in the math on which the transformer is based there is no restriction on the length of input. But I still can't understand why we should fix it in the frameworks (PyTorch). Because of this problem Transformer-XL has been created.

Can you explain to me where this problem is hiding, please?"

Answer:
"The restriction in the maximum length of the transformer input is due to the needed amount of memory to compute the self-attention over it.

The amount of memory needed by the self-attention in the Transformer is quadratic on the length of the input. This means that increasing the maximum length of the input, increases drastically the needed memory for self-attention. The maximum length is that which makes the model use up the whole memory of the GPU for at least one sentence (once the other elements of the model are also taken into account, like the embeddings which take a lot of memory).

Transformer-XL is certainly a way to take into account as much context as possible in language modeling (its role is analogous to truncated back-propagation through time in LSTM language models). However, the gradients are not propagated through the attention over the memory segment, only through the current segment.

There have been several architectural attempts to reduce the amount of memory needed by transformers, like using locality-constraints in the attention (Dynamic Convolutions model) or using locality-sensitive hashing (Reformer model).

There have been other implementation attempts, like gradient checkpointing(e.g. this), which is a general technique to run computations that don't fit at once in the GPU memory"
\end{lstlisting}

\LaTeX{} formatting refinement:
\begin{lstlisting}[style=prompt]
You will be given a text. Your only task is to correct any mathematical notation inside it to be valid LaTeX. You must not change any other part of the text.
    - Convert unicode math characters like 'pi' to their LaTeX equivalent '$\\pi$'.
    - Ensure all mathematical expressions are enclosed in '$...$' for inline math or '$$...$$' for display math.
    - Return the full, corrected text.
\end{lstlisting}

\textbf{2. Prompt to synthesize textbook-style explanations:}

Textbook outline generation:
\begin{lstlisting}[style=prompt]
### Instructions
You will be given a research paper and your task is to create a detailed outline for a textbook that comprehensively explains the given research paper. But, it should go beyond mere explaining, and be a proper pedagogical textbook that aims to fully educate the reader on what the paper is about. The textbook should be aimed at college students who have a basic understanding of machine learning.

The outline should:
- Break down the paper into coherent chapters.
- For each chapter, provide a:
    - title
    - description
    - list of subtopics to cover
- Cover all key concepts, methods, and results from the paper.
- Ensure a logical flow of information, from introduction to conclusion.
- While the textbook should be comprehensive, it should also articulate and to the point. Don't create unnecessary chapters.

### Output Format
Provide the output as a JSON object with a single key "outline", which is a list of chapter objects. Each chapter object must have the following keys:
- "chapter_title": A string for the title of the chapter.
- "description": A string describing the chapter's content.
- "subtopics": A list of strings, where each string is a subtopic.
\end{lstlisting}

Chapter generation:
\begin{lstlisting}[style=prompt]
### Instructions
You will be given a chapter title, description, and subtopics and, based on those topics, your job is to write a detailed, cohesive textbook chapter addressed to a college student who is learning this material for the first time. 

The chapter should be comprehensive and suitable for someone learning this material to understand research papers in the field. Don't just briefly describe the subtopics, but rather elaborate on the concepts at full length and explain them with a focus on intuition. Spell everything out clearly so there is no ambiguity. Dedicate multiple paragraphs to each subtopic but be articulate and concise when appropriate. Write in full prose, rather than bullet points. Most importantly, please make sure that your chapter is grounded in the paper; do not provide any information or details that is not from the paper.

Start with the chapter title in the first line. Separate each subtopic with a section header "#". Also, please write all mathematical notation in LaTeX only e.g. "$x^2$" or "$\pi$". Do not use unicode mathematical characters e.g. "pi". Again, PLEASE write all math in LaTeX.
\end{lstlisting}

\textbf{3. Prompt to synthesize blog-post-style explanations:}

Blog post idea generation:
\begin{lstlisting}[style=prompt]
### Instructions
You are a creative tech blogger and content strategist. Based on the provided research paper, generate a list of a few blog posts that explain the paper in a way that is accessible to a wider audience. They should each focus on a different, main aspect of the paper.

For each blog idea, provide:
- A `title`.
- A brief `description` of what the blog post will cover.

### Output Format
Provide the output as a JSON object with a single key "blogs", which is a list of blog objects. Each blog object must have the following keys:
- "title": A string for the title of the blog post.
- "description": A string describing the blog post's content.
\end{lstlisting}

Blog post generation:
\begin{lstlisting}[style=prompt]
You will be given an academic paper and a blog post idea about the paper. Write a blog post based on the blog idea.

As you write the blog post, please make sure to consider the following:
- Write in a technical blog style. It should be less formal but not too informal. It should be concise and to the point. 
- Simplify complex concepts from the paper for a broader audience.
- Write in full, complete sentences and prefer paragraphs over bullet points, but use bullet points when appropriate.
- Keep all details grounded in the paper. Do not make up any information.
- Please write any mathematical notation in LaTeX only e.g. "$x^2$" or "$\pi$". Do not use unicode mathematical characters e.g. "pi". 

Your output should be the full text of the blog post, starting with the blog title as a markdown header. Use '#' to denote the blog title, '##' to denote different sections, and so on.
\end{lstlisting}

\textbf{4. Prompt to generate prerequisite-knowledge chapters:}

Prerequisite chapter-list generation:
\begin{lstlisting}[style=prompt]
### Instructions
You are an expert curriculum designer. Based on the provided research paper, create a list of textbook chapters that would provide all the necessary prior knowledge to understand this paper. The chapters should not contain the novel ideas presented in the paper itself, but rather the foundational concepts upon which the paper is built.

For each chapter, provide:
- A `title`.
- A general `description` of what the chapter covers.
- A list of `subtopics` that should be included.

### Output Format
Provide the output as a JSON object with a single key "chapters", which is a list of chapter dictionaries.
Example:
{
  "chapters": [
    {
      "title": "Chapter 1: Introduction to Probability Theory",
      "description": "This chapter covers the basics of probability...",
      "subtopics": ["Random Variables", "Probability Distributions", "Bayes' Theorem"]
    }
  ]
}
\end{lstlisting}

Chapter generation:
\begin{lstlisting}[style=prompt]
### Instructions
You will be given a chapter title, description, and subtopics and, based on those topics, your job is to write a detailed, cohesive textbook chapter addressed to a college student who is learning this material for the first time. 

The chapter should be comprehensive and suitable for someone learning this material to understand research papers in the field. Begin with an introduction to the chapter, then cover each subtopic in turn. Don't just briefly describe the subtopics, but rather elaborate on the concepts at full length and explain them with a focus on intuition. Spell everything out clearly so there is no ambiguity. Dedicate multiple paragraphs to each subtopic. Write in full prose, rather than bullet points. 

Separate each subtopic with a section header "#".

Also, please write all mathematical notation in LaTeX only e.g. "$x^2$" or "$\pi$". Do not use unicode mathematical characters e.g. "pi".
\end{lstlisting}

\end{document}